\documentclass[10pt,twoside]{amundi_article}

\usepackage[table]{xcolor}
\usepackage{lmodern}
\usepackage{amssymb}
\usepackage{amsfonts}
\usepackage{amsmath}
\usepackage{amsbsy}
\usepackage{fancybox}
\usepackage{fancyhdr}
\usepackage{graphicx}
\usepackage[hyphens]{url}
\usepackage{arydshln}
\usepackage{multirow}
\usepackage{pdflscape}
\usepackage{tikz}
\usepackage{comment}
\usepackage{nicefrac}
\usepackage{dsfont}
\usepackage{algorithmic}
\usepackage{algorithm}
\usepackage{lipsum}

\newlength{\figurewidth}
\newlength{\figureheight}
\DeclareMathAlphabet\mathbfcal{OMS}{cmsy}{b}{n}
\DeclareFontFamily{OT1}{pzc}{}
\DeclareFontShape{OT1}{pzc}{m}{it}{<-> s * [1.3] pzcmi7t}{}
\DeclareMathAlphabet{\mathpzc}{OT1}{pzc}{m}{it}

\usetikzlibrary{positioning}
\usetikzlibrary{patterns,arrows,decorations.pathreplacing}
\usetikzlibrary{backgrounds}

\usepackage[
        left=3.0cm,
        right=3.0cm,
]{geometry}

\newlength\FHoffset
\fancyheadoffset{\FHoffset}

\usepackage[authoryear]{natbib}

\usepackage{ragged2e}

\usepackage{subfigure}
\usepackage{booktabs}

\definecolor{amundi_blue}{RGB}{0,176,240}
\definecolor{amundi_dark_blue}{RGB}{0,28,75}
\definecolor{darkblue}{rgb}{0.0, 0.0, 0.55}
\usepackage{hyperref}
\hypersetup{
    colorlinks=true,
    citecolor=darkblue,
    linkcolor=darkblue,
    filecolor=darkblue,      
    urlcolor=darkblue
}

\usepackage{bookmark}

\usepackage{enumitem}
\usepackage{balance}

\usepackage{tabularx}

\begin{document}

\setcounter{page}{1}

\title{\textbf{\color{amundi_blue}Encoded Value-at-Risk: \\
A Predictive Machine for Financial Risk Management}%
}

\author{
\hspace{-0.3cm} {\color{amundi_dark_blue} Hamidreza Arian}\footnote{E-Mail: {\color{amundi_blue} \href{emailto:hamidreza.arian@utoronto.ca}{hamidreza.arian@utoronto.ca}}} \\
\hspace{-0.3cm} Sharif University \\
\hspace{-0.3cm} of Technology
\and
{\color{amundi_dark_blue} Mehrdad Moghimi}\footnote{E-Mail: {\color{amundi_blue} \href{emailto:mehrdad.m7496@gmail.com}{mehrdad.m7496@gmail.com}}} \\
Sharif University \\
of Technology
\and
{\color{amundi_dark_blue} Ehsan Tabatabaei}\footnote{E-Mail: {\color{amundi_blue} \href{emailto:ehsan.taba94@gmail.com}{ehsan.taba94@gmail.com}}} \\
Khatam \\
University
\and
{\color{amundi_dark_blue} Shiva Zamani}\footnote{E-Mail: {\color{amundi_blue} \href{emailto:zamani@sharif.edu}{zamani@sharif.edu}}} \\
Sharif University \\
of Technology
}

\date{\color{amundi_dark_blue}August 2020}

\maketitle

\begin{abstract}
\noindent
Measuring risk is at the center of modern financial risk management. As the world economy is becoming more complex and standard modeling assumptions are violated, the advanced artificial intelligence solutions may provide the right tools to analyze the global market. In this paper, we provide a novel approach for measuring market risk called Encoded Value-at-Risk (Encoded VaR), which is based on a type of artificial neural network, called Variational Auto-encoders (VAEs). Encoded VaR is a generative model which can be used to reproduce market scenarios from a range of historical cross-sectional stock returns, while increasing the signal-to-noise ratio present in the financial data, and learning the dependency structure of the market without any assumptions about the joint distribution of stock returns. We compare Encoded VaR out-of-sample results with eleven other methods and show that it is competitive to many other well-known VaR algorithms presented in the literature.  
\end{abstract}

 \medskip


\section{Introduction}\label{sec:intro}

Analysis of fat-tailed distributions and calculating an appropriate risk measure have always posed great challenges to risk managers and practitioners in financial markets. One of the most popular measures for  market  risk is Value-at-Risk (VaR), which was proposed by JP. Morgan in 1990.  There are four main methods for calculating VaR; parametric, semi-parametric, non- parametric and hybrid approaches. 
Substantial efforts have been directed to relax the limitations of these approaches and to provide a more realistic prediction of market risk. These limitations include normality and linear dependency assumptions on return time series, as well as the computational expense one should deal with when forecasting risk. As we will discuss below, there have been many attempts to fix some of these weaknesses and propose new approaches to capture fat-tailed distributions, asymmetry, and non-linear dependency in the financial markets.

It is a stylized fact that the returns of financial assets are fat-tailed (\cite{Mandelbrot2002ThePrices}). One of the popular approaches to model fat-tail downside risk of financial assets uses Extreme Value Theory (EVT), which  models only the tail of the return distribution. However, EVT relies also on an extreme risk threshold which should be chosen delicately, thereby making its usage challenging. \cite{Lin2009PortfolioTheory} propose a genetic algorithm (GA) approach  to estimate an optimal threshold  for a portfolio. They apply their method to a portfolio of 78 stocks listed on the Taiwan stock exchange and conclude that the VaR estimated by GA-based EVT method provides acceptable results. Due to the popularity of the EVT approach, it has been combined with other predictive routines. \cite{McNeil2000EstimationApproach} propose a method for estimating VaR using a hybrid of EVT and GARCH volatility models. They use data over the period of January 1960 to June 1993 and show that their approach very well captures the fat-tail property of return distribution as well as the conditional volatility.\cite{CHANG2007ForecastingMethod} propose a different approach for measuring Value-at-Risk. Their Percentile of Cluster Method, instead of estimating the distribution, separates data into clusters and ranks them to calculate VaR from extreme clusters.

Asymmetric features pose another challenge for modeling and predicting market risk. 
\cite{Levich1985EmpiricalEfficiency} provides evidence that the volatility of financial assets are asymmetric. In general, using volatility for risk forecasting needs to be done carefully as option-based implied volatility measures tend to under-perform when compared to their historical counterparts (\cite{Bams2017}), while choosing the loss function may change this comparative result (\cite{Gonzalez-Rivera2004}). Considering both fat-tail properties and asymmetric volatililities, \cite{Gencay2004ExtremeMarkets} combine the EVT approach with asymmetric volatility models and calculate VaR. They compare the estimated parameters in developing and developed countries and report that the parameters of their model differ significantly between the two subgroups. \cite{Taylor2008EstimatingExpectiles} estimates VaR using linear asymmetric least squares regression (LALSR). Their approach has shown promising results although  not considering non-normality and non-linear features. In \cite{Wang2011MeasuringRegression}, kernel method is used to add non-linearity to LALSR method. 

For building a robust predictive risk model, in addition to tail and asymmetric behaviour, the relationship between financial assets should be taken into account. Copula functions may be used to model the dependency structure in the market (\cite{Karmakar2019}). \cite{Wang2010EstimatingMarket} estimate VaR of a foreign exchange portfolio using a GARCH-EVT-Copula model. Then, they solve an asset-allocation problem and find out that t Copula and Clayton Copula, are more accurate in modeling the correlation structure of the assets in a portfolio.

\cite{Engle2004CAViaR:Quantiles} propose an approach called Conditional Autoregressive VaR by regression quantiles (CAViaR) using an autoregressive process for VaR itself in both symmetric and asymmetric regression setups on various types of factors. Their approach has been followed by many improvements including an extension accounting for intra-day price ranges (\cite{Chen2012}), addressing trading-loss limits (\cite{Fuertes2013}) and overnight returns and close-to-close ranges (\cite{Meng2018}). 

Other than fat-tail properties of the return distributions, the long-memory effect of the volatility is studied in \cite{Youssef2015Value-at-RiskApproach}, with a long-memory GARCH EVT approach for the energy market. Their findings show that using long-term memory asymmetry and fat-tail positively affects the risk management and profits of hedging strategies. Other challenges in building risk measurement algorithms are due to market's regime changes (\cite{Ardia2018}), seasonal behaviour (\cite{FongChan2006}) and time-varying volatility, kurtosis and skewness (\cite{Guermat2002}, \cite{Lucas2016}). 

\begin{figure*}
    \centering
    \subfigure[MNIST 2-D latent space]{{\includegraphics[width=8.3cm]{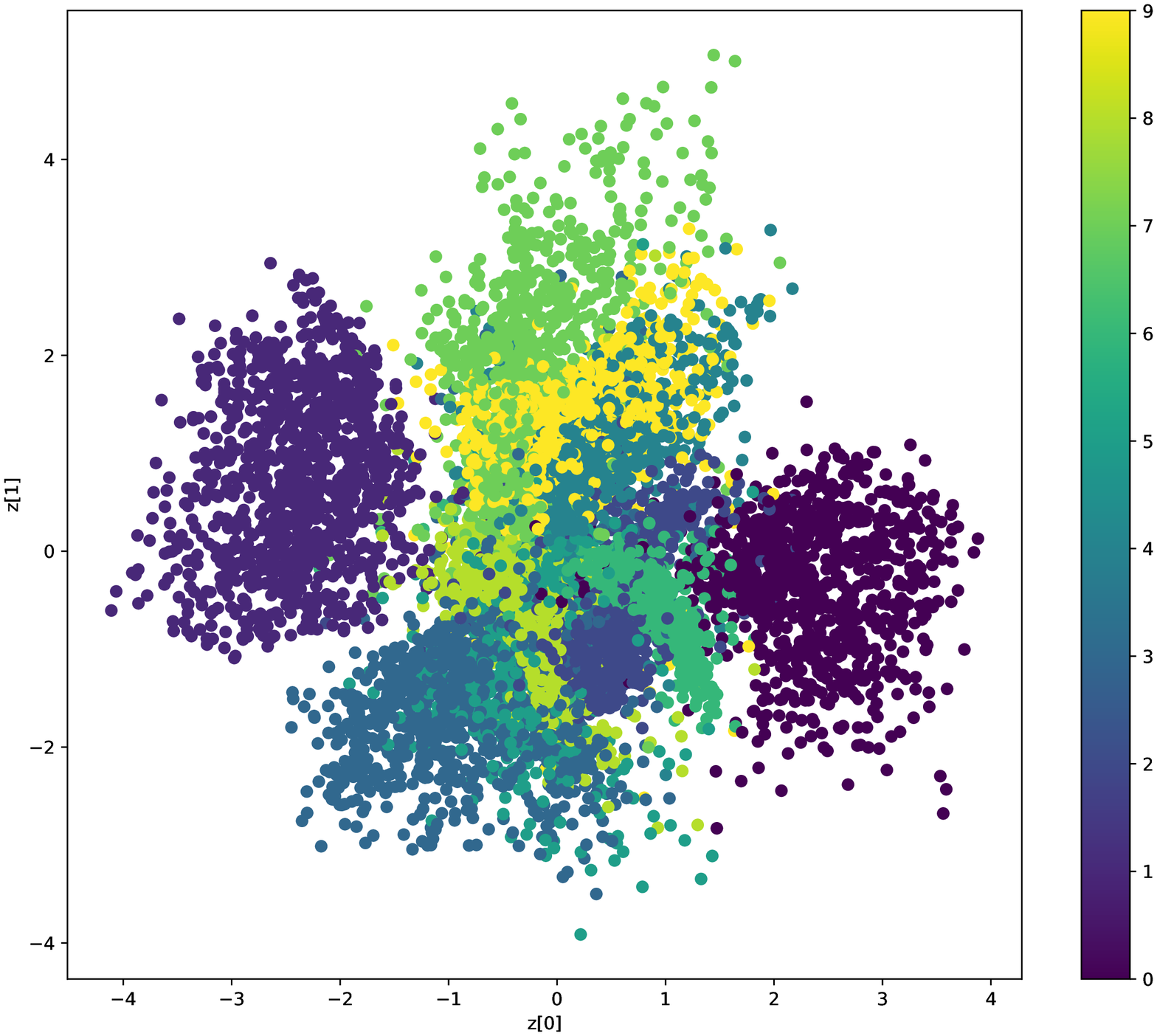}}}
    \subfigure[2-D latent space MNIST manifold]{{\includegraphics[width=7.0cm]{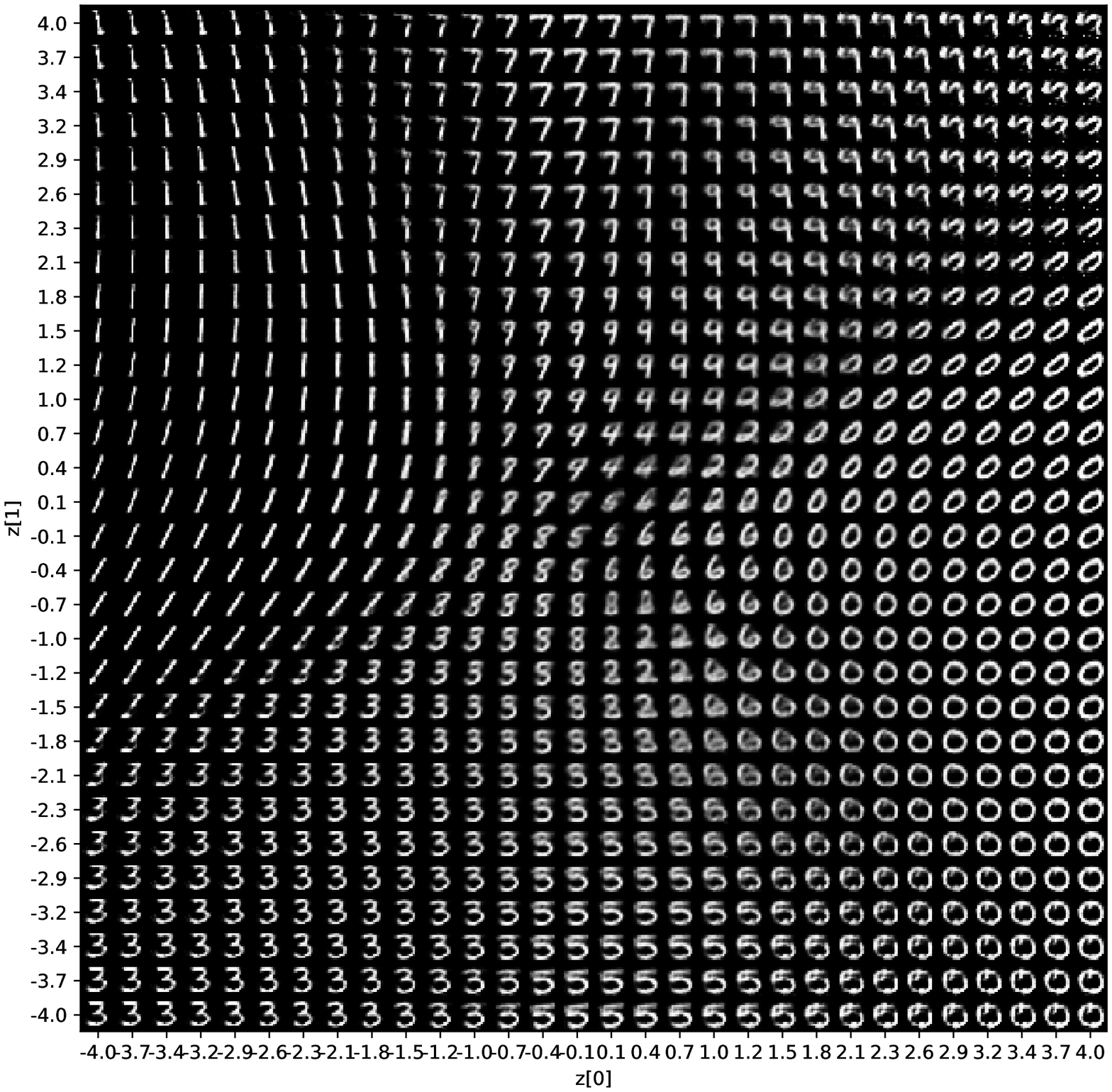} }}
    \caption{Latent space of a trained Variational Auto-encoder on MNIST data (panel a) and its generated output manifold (panel b).}%
    \label{fig:vae_mnist}%
\end{figure*}

Using neural network models have been quite popular among researchers in recent years. One of the reasons is having few assumptions on the modeling side when compared to other conventional approaches. As \cite{LopezdePrado2019} suggests, a generative neural network can be used for financial data set augmentation. When this model learns the statistical behaviour of its input data, it can generate new observations with the same properties. We discuss more about this model in the following sections.

Using a generative neural network, one can let the model learn the shape and dependency structure of asset returns, and then the model can generate many scenarios which are likely to occur in the future. Auto-encoders are one of such neural networks which were originally used in image processing (\cite{Doersch2016TutorialAutoencoders}). These generative models can reproduce new pictures similar to the ones which were used to train the model. An auto-encoder consists of two neural networks called encoder and decoder. The encoder takes an input and converts it to a smaller, dense representation. The decoder then tries to convert this compact representation back to the original input. These two networks operate consecutively and are trained together. The latent variable must have enough information so that the decoder is able to process it into the desired output format, and because the latent space has much lower dimension, the encoder must learn to preserve the relevant information in the input. When the model is trained to reconstruct its own input, the loss function is usually either the mean-squared error or cross-entropy between the input and the output. With these loss functions, the model is penalized when it creates outputs that are different from the input. Auto-encoders do not impose any restrictions on the latent space, therefore it might be discontinuous. In such situation, taking random samples from such latent space and feeding it to the decoder could generate data which are not statistically similar to our training data. To solve this problem,  \cite{Kingma2014AutoEncodingBayes} propose a method which led to the development of Variational Auto-Encoders (VAEs). 
VAEs take advantage of variational inference to impose restrictions on the latent space and make it close to a known distribution. This way, the latent space can be sampled and fed to the decoder to generate new data.

For illustration, we have trained a VAE model with MNIST Dataset \footnote{Available at \url{http://yann.lecun.com/exdb/mnist}} with 2-D latent space, as suggested in  \cite{Kingma2014AutoEncodingBayes}. Figure \ref{fig:vae_mnist} panel (a) shows the latent mean values on digit numbers of MNIST Dataset. As it can be seen, digit numbers are almost separable having just the means of the latent space. By assigning the means of the latent space, we can choose which digit we want the model to generate. In figure \ref{fig:vae_mnist} panel (b), the means of the latent space are gradually changed in the interval $[-4,4]$ and the generated pictures are shown. As it is clear from Figure \ref{fig:vae_mnist} panel (b), one can see that the digits  transform smoothly from one to another.


In this paper, we provide a novel approach, which we call Encoded Value-at-Risk (Encoded VaR). In this approach, we estimate various quantiles of an unknown distribution using Variational Auto-encoders. After learning the shape of the return distribution, the model can generate daily returns that statistically resemble our training data.

Using our normalization method, Encoded VaR imposes regime changes and seasonality effects automatically and therefore can quickly realize when to increase (decrease) the risk forecasts while approaching (getting out of) a financial crisis. Our method can learn the shape of return distributions, and therefore  is able to estimate VaR without making any distributional assumptions. Due to the fact that non-linear activation functions are incorporated into neural networks, the model is also able to learn both linear and non-linear dependencies of the assets in the market. Therefore, there is no more need to use copula functions or any other measure of dependency at all.


In what follows, section \ref{sec:method} provides an overview of Value-at-Risk for measuring the risk of financial portfolios, and the Variational Auto-encoder as a powerful generative model. We then introduce our Encoded VaR methodology and show how it can be used to generate scenarios for forecasting VaR. In section \ref{sec:emp}, we present numerical results on three portfolios from major exchanges and compare the performance of the Encoded VaR with several other powerful techniques. Section \ref{sec:con} concludes the paper and lays down future research directions about risk forecasting using neural networks.

\section{Methodology}\label{sec:method}

In this section, we give our proposed method for estimating Value-at-Risk using neural networks simulation. First, we will have a brief review on VaR, and then we provide the conceptual modeling framework for measuring the risk of investment portfolios with Variational Auto-encoders. 

\subsection{Measuring Financial Risk}

Value-at-Risk (VaR) is a popular measure of risk defined on a given horizon with a confidence level $1-\alpha$, such that loss beyond VaR occurs with probability $\alpha$. The main advantages of VaR are its simplicity and economic intuition. We define  $x_t=\log(\frac{P_t}{P_{t-1}})$ as the daily logarithmic return. Let $X$ represent the distribution of these daily returns, then $\text{VaR}_{\alpha}(X)$ will be the $\alpha$-quantile of $X$
\begin{equation}
\label{eq:var}
    \text{VaR}_{\alpha}(X) = \sup \left\{x \in \mathbb{R}: F_{X}(x)< \alpha\right\}=F_{X}^{-1}(\alpha),
\end{equation}
where $F_X$ is the cumulative distribution function (CDF) of X. Although VaR  is usually defined as a positive number, for simplicity we consider it negative as seen in \ref{eq:var}. There are many approaches to estimate VaR, including parametric, non-parametric, semi-parametric, and hybrid methods. Here, we propose the Encoded VaR approach which belongs to the semi-parametric category. 

\begin{figure}
    \centering
    {\includegraphics[width=16.0cm]{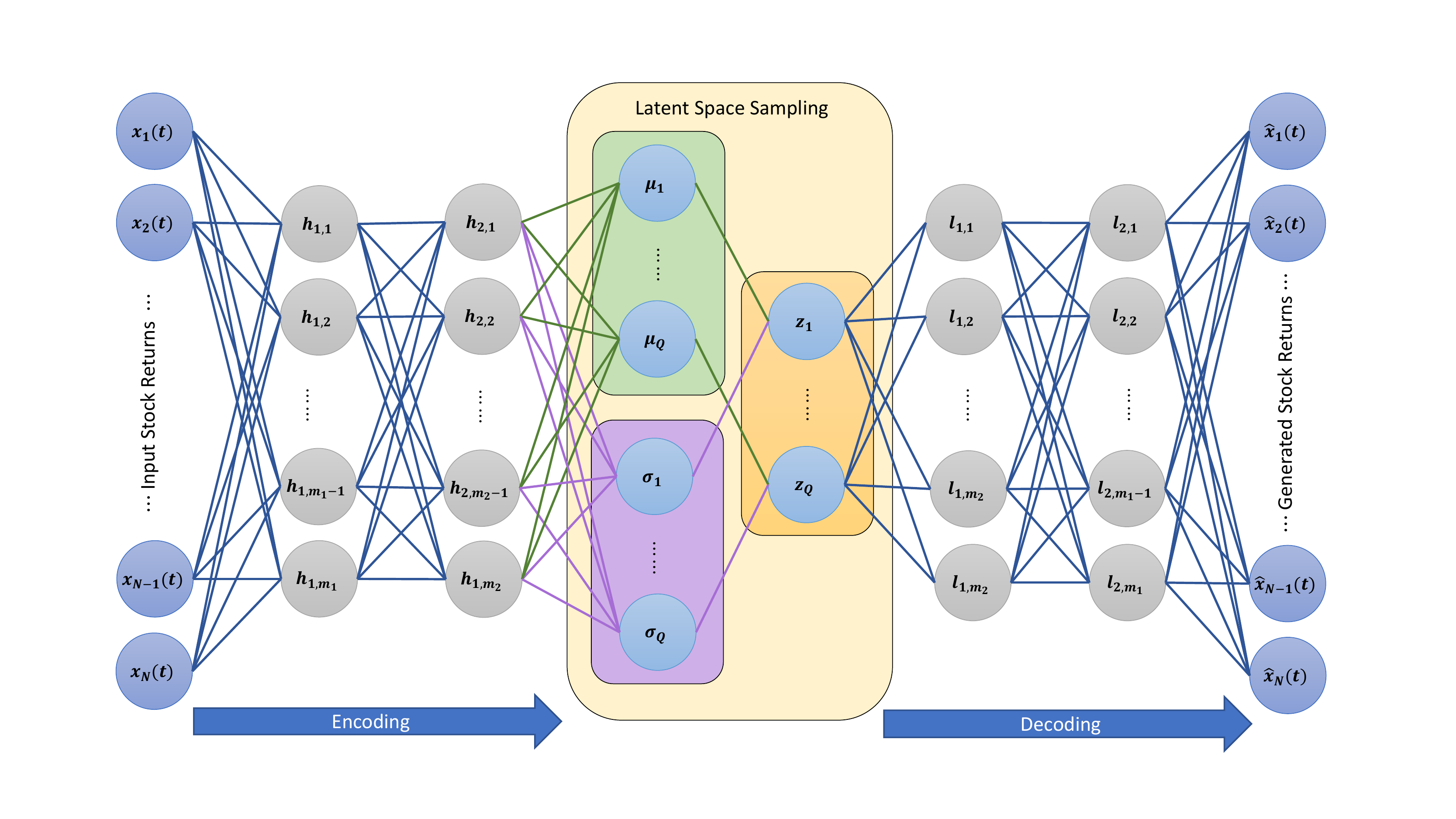}}
    \caption{The general structure of a VAE. The left (right) portion shows the encoder (decoder).}
    \label{fig:vae_diagram}
\end{figure}

\subsection{Variational Auto-Encoder}

Let us assume $x$ represents a sample from real market data and $z$ is a sample variable in the latent space. The VAE consists of a probabilistic encoder model  $q_{\phi}(z|x)$ and a probabilistic decoder model $p_{\theta}(x|z)$, given a fixed prior $p_{\theta}(z)$, where $\phi$ and $\theta$ are the neural network weights. Here, $q_{\phi}(z|x)$ maps the real market data space to the latent space, where for each sample, $x$, a latent variable, $z$, in a lower dimensional space is assigned. For generating new data, first, a value $z^{(i)}$ is sampled from the prior distribution $p_{\theta}(z)$, then $x^{(i)}$ is generated using the conditional distribution $p_{\theta}(x|z)$. Our goal is to find the true posterior $p_{\theta}(z|x)$
\begin{equation}
    p_{\theta}(z|x) = \dfrac{p_{\theta}(x|z)p_{\theta}(z)}{p_{\theta}(x)} = \dfrac{p_{\theta}(x|z)p_{\theta}(z)}{\int{ p_{\theta}(x|z) p_{\theta}(z) dz}}. 
    \label{eq:Bayes}
\end{equation}
However, the marginal likelihood $p_{\theta}(x)$ is intractable, and therefore, it cannot be evaluated and differentiated with respect to $\theta$. To solve this problem, an arbitrary distribution $q(z)$ can be used to approximate the true posterior $p_{\theta}(z|x)$. We do this by minimizing the Kullback-Leibler (KL) divergence between $q(z)$  and $p_{\theta}(z|x)$
\begin{equation}
q(z)=\arg \min _{q} \mathcal{KL}(q(z)||p_{\theta}(z|x)).
\end{equation}
The KL divergence is a measure defined on probability distributions as 
\begin{equation} 
\label{eq:2.3}
    \mathcal{KL}(q(z)||p_{\theta}(z|x)) = \sum  q(z) \log \frac{q(z)}{p_{\theta}(z|x)} = \mathbb{E}_{q(z)}\left[\ln \frac{q(z)}{p_{\theta}(z | y)}\right].
\end{equation}

Rewriting the equation \ref{eq:2.3} shows that solving this optimization problem also involves computing the marginal likelihood
\begin{equation}\label{eq:2.4}
\begin{aligned}
   \mathcal{KL}(q(z)||p_{\theta}(z|x)) = & \mathbb{ E}_{q(z)}\left[\ln \frac{q(z)}{p_{\theta}(z | x)}\right] \\ 
   = &\mathbb{ E}_{q(z)}[\ln q(z)]-\mathbb{E}_{q(z)}[\ln p_{\theta}(z | x)]\\  
   = & \mathbb{ E}_{q(z)}[\ln q(z)]-\mathbb{E}_{q(z)}[\ln p_{\theta}(z, x)]+\ln p_{\theta}(x).
\end{aligned}
\end{equation}
By re-arranging  equation \ref{eq:2.4} , It can be seen that the constant value $\log p_{\theta}(x)$ is the summation of two term. We want to minimize on of these terms, but  cannot evaluate it, so we maximize the other term. Knowing that $\mathcal{KL}(q_{\phi}(z)||p_{\theta}(z|x)) \geq 0$, we can define the lower bound as 
\begin{equation}
\begin{aligned}
\ln p_{\theta}(x) = &  \mathcal{KL} (q(z)||p_{\theta}(z | x)) \\
+ & \Big(-\mathbb{E}_{q(z)}[\ln q(z)]+\mathbb{E}_{q(z)}[\ln p_{\theta}(z, x)]\Big) \\
\geq &  -\mathbb{ E}_{q(z)}[\ln q(z)]+\mathbb{E}_{q(z)}[\ln p_{\theta}(z, x)] \\
= &  -\mathbb{ E}_{q(z)}[\ln q(z)]+\mathbb{E}_{q(z)}[\ln p(z)]+\mathbb{E}_{q(z)}[\ln p_{\theta}(x | z)] \\
= &  -\mathcal{KL}(q(z) | p_{\theta}(z))+\mathbb{E}_{q(z)}[\ln p_{\theta}(x | z)].
\end{aligned}
\end{equation}
Although the model distribution $q(z)$ can be any arbitrary distribution, we assume it depends on $x$, and use the encoder model $q_{\phi}(z|x)$, as we want to reconstruct any input $x$ from its corresponding latent variable $z$ in the decoder. Therefore, the model can be trained by maximizing the lower bound defined as
\begin{equation}
\label{eq:lowerbound}
\mathcal{L}(\theta, \phi) = -\mathcal{KL}(q_{\phi}(z|x) | p_{\theta}(z))+\mathbb{E}_{q_{\phi}(z|x)}[\ln p_{\theta}(x | z)].
\end{equation}
 If we assume that both $q_{\phi}(z|x)$ and $p_{\theta}(z)$ have Gaussian distributions, we have 
\begin{equation}
\begin{aligned}
q_{\phi}(z|x) =& \mathcal{N}\Big(z ; \mu(x ; \phi), \operatorname{diag}(\sigma^{2}(x ; \phi))\Big), \\
p_{\theta}(z) =& \mathcal{N}(z ; 0, I_{Q}).
\end{aligned}
\end{equation}
where $Q$ is the dimension of the latent space, $I_{Q} \in \mathbb{R}^{Q \times Q}$ denotes the identity matrix and $\mathcal{N}$ refers to the Gaussian distribution. The KL divergence of two Gaussian distributions has a closed form which is given in \cite{Kingma2014AutoEncodingBayes}. Therefore, the first part of the lower bound can be written as
\begin{equation}
\label{eq:klclosed}
\mathcal{KL} (q_{\phi}(z|x) | p_{\theta}(z))= \frac{1}{2} \sum_{i=1}^{Q} \Big(- \ln \sigma_{i}(x)+ \sigma_{i}(x)^{2}+ \mu_{i}(x)^{2}-1\Big).
\end{equation}

Since the decoder in our model is deterministic, it assigns only one value ($\hat x$) to any given latent variable $z$. Thus, we can write $p_{\theta}(x | \hat x)$ instead of $p_{\theta}(x | z)$. Supposing that our input (and output) data follow a normal distribution, we can replace $p_{\theta}(x | \hat x)$ by its distributional form, and $\ln p_{\theta}(x | \hat x)$ can be written as the mean squared difference of input and output. It is worth noting that we only use the normality assumption in the parameter optimization of the model, and VaR calculation still has no assumptions about the distribution of the input data.

The negative of the lower bound can be considered as the loss function of the model that should be minimized.  With the aforementioned equation, we saw that both terms of this loss function can be calculated. 

One last remark on the implementation of the VAE, is that the latent layer needs a simple reparameterization trick to be differentiable w.r.t $\phi$ in the model optimization. As discussed earlier, we randomly sample from a multivariate normal distribution at the latent layer and feed them to the decoder side of the network. This distribution has mean, $\mu$, and standard deviation, $\sigma$, derived from the encoder network. Instead, we randomly sample from a multivariate standard normal distribution, $\epsilon$. Afterwards, by adding and multiplying $\mu$ and $\sigma$, respectively, we calculate $z$ as $z = \mu + \sigma\epsilon$ (see~\cite{Kingma2014AutoEncodingBayes}).

\subsection{Encoded VaR}

The proposed method in this article, named Encoded-VaR, uses a VAE to generate new samples of standardized returns. The architecture of the VAE has the ability to learn the dependency structure and other statistical features in our input data, and its controlled latent space makes sampling possible. Consider we have $N$ risky assets $S_1$, $S_2, \ldots, S_N$, with time series of returns $x_{1}, x_{2}, \ldots, x_{N}$. The return of a portfolio $\mathcal{P}$, with weights $\boldsymbol{\Omega} = (\omega_1, \ldots, \omega_N)$, at time $t$ is

\begin{equation}
r_\mathcal{P}(t; \boldsymbol{\Omega}) = \sum_{i=1}^N{\omega_i {x}_{i}(t) }.
\end{equation}
Here, we represent the steps to estimate VaR.

\subsubsection{Data preprocessing and standardization}
Deep learning models require standardized input to improve the numerical stability of the model and to reduce the training time.  We standardize each daily return according to its rolling mean and standard deviation, $\mu_{i,t,w}$ and $\sigma_{i,t,w}$, calculated from the window $[x_{i}(t-w),x_{i}(t-1)]$, where $w$ is the size of the window. The standardized return, ${x}^{\mathcal{S}}_{i}(t)$, is calculated as 
\begin{equation}
\label{eq:var}
    {x}^{\mathcal{S}}_{i}(t) = \frac{x_{i}(t) - \mu_{i,t,w}}{\sigma_{i,t,w}} \qquad i=1, 2, \ldots, N.
\end{equation}
\subsubsection{Training the model}
After normalizing the data, the model is trained and its parameters are estimated.  If the generated data by the VAE are shown by $\hat{x}^{\mathcal{S}}_{1}(t)$, $\hat{x}^{\mathcal{S}}_{2}(t)$, $\ldots$, $\hat{x}^{\mathcal{S}}_{N}(t)$, according to equations \ref{eq:lowerbound} and \ref{eq:klclosed}, the loss function has the form
\begin{equation}
\begin{aligned}
\mathcal{L}(\phi, \theta; \mathbf{x}^{\mathcal{S}}) & =C\sum_{i=1}^N {\Big({x}^{\mathcal{S}}_{i}(t)-\hat{{x}}^{\mathcal{S}}_{i}(t)\Big)^{2} }\\
+& \frac{1}{2} \sum_{j=1}^{Q} \Big(- \ln \sigma_{j}({x}^{\mathcal{S}}(t))+ \sigma_{j}({x}^{\mathcal{S}}(t))^{2}+ \mu_{j}({x}^{\mathcal{S}}(t))^{2}-1\Big)    
\end{aligned}
\end{equation}
where $\phi$ and $\theta$ are encoder and decoder weights, respectively,  $C$ is a regularizer and time $t$ is in our training time-frame. This loss function is minimized using a Stochastic Gradient Descent algorithm. After the model is trained, it can be used to generate new samples similar to the input data.

\subsubsection{Sampling from latent space and the decoding}
After training, the distribution of bottleneck layer asymptotically approaches to a multivariate normal distribution. Now, for each test day, $s$, we can generate thousands of samples, $\zeta$, from the latent space, where each of them can be mapped to a scenario for daily stock returns. The flexibility of the Encoded VaR allows us to sample from normal distribution while the regenerated stock return samples have an unknown distribution. Then, we use the decoder side of the network, represented by the function $\mathcal{D}$, to produce the output returns $\hat{\mathbf{x}}^{\mathcal{S}}_{s} = (\hat{{x}}^{\mathcal{S}}_{1}(s), \hat{{x}}^{\mathcal{S}}_{2}(s), \ldots, \hat{{x}}^{\mathcal{S}}_{N}(s)) $ by
\begin{equation}
\label{eq:var}
    \hat{\mathbf{x}}^{\mathcal{S}}_{s} = \mathcal{D}(\zeta).
\end{equation}
The generated samples $\hat{\mathbf{x}}^{\mathcal{S}}_{s} $ do not carry any information regarding the mean and the standard deviation of real data, and therefore, need to be destandardized before calculating VaR.  

\subsubsection{Destandardizing the data}
We will use the mean and standard deviation of the last rolling window $[x_{i}(s-v), x_{i}(s-1)]$ to destandardize the generated returns, where $v$ is the size of the window, and $s$ is in our testing time-frame. This destandardization method can be considered as a way to impose the predicted mean and standard deviation for $\hat{{x}}^{\mathcal{S}}_{i}(s)$ by
\begin{equation}
\label{eq:var}
    \hat{x}^{\mathcal{D}}_{i}(s) = \sigma_{i,s,v}.\hat{{x}}^{\mathcal{S}}_{i}(s) + \mu_{i,s,v}.
\end{equation}
Using standardized returns with rolling windows in our model serves one important purpose. Financial time series often have dynamic variance, and normalizing with a rolling window makes the distribution of input data almost identical. Moreover, in the denormalization step, the rolling windows can be used to impose the latest predicted mean and standard deviation on the output of the VAE. In this article, we use exponentially weighted moving average and standard deviation to normalize and denormalize the returns. However, the Encoded VaR methodology can also be equipped with a wide range of predictive algorithms for the return and standard deviation, such as ARIMA and GARCH.

\subsubsection{Calculating VaR}
Now we can use the generated scaled returns, $\hat{x}^{\mathcal{D}}_{i}(s)$, to calculate the return of portfolio $\mathcal{P}$
\begin{equation}
r_\mathcal{P}^\mathcal{D}(s; \boldsymbol{\Omega}) = \sum_{i=1}^N{\omega_i \hat{x}^{\mathcal{D}}_{i}(s) }.
\end{equation}
By regenerating, we can reconstruct the distribution of the portfolio return and calculate VaR as the $\alpha\text{th}$ percentile of returns. It is a stylized fact that the noise-to-signal ratio is high in the financial data sets (\cite{prado_2018}, \cite{prado_2020}). This is a major reason that most of the financial models are unstable and provide sensitive outputs with respect to slight changes in the market environment. Our Encoded VaR algorithm is built on an auto-encoder network, which uses dimension reduction and reconstruction. Therefore, it decreases the noise-to-signal ratio significantly by re-sampling from the latent space . In section \ref{sec:emp}, we will see that our generative model reduces the \textit{noise}, while retaining the \textit{relevant} information in the data. 

We can see that in the Encoded VaR engine, not only the model can reconstruct the input data, but also we have control over the latent space. Therefore, the decoder, when trained, can be used to generate new samples of stock returns. Moreover, the data is fed into the model in a way that the neurons on the input layer are cross-sectional and the samples are time series.  A general diagram showing the VAE's architecture is presented in Fig. \ref{fig:vae_diagram}.
\section{Empirical Results}\label{sec:emp}

In this section, we first introduce the loss functions which are used for backtesting our model. Then we provide some information with regards to our input data and the corresponding Encoded VaR results. Next, we compare our model with eleven other well-investigated Value-at-Risk models proposed in the literature. We use Python and TensorFlow for implementing our model. For backtesting, we compare the Encoded VaR approach with benchmark models using multiple loss functions like Sener’s Penalization Measure, Quantile, Lopez, Sarma, Linear, and Quadratic loss functions.

\subsection{Loss Functions}
Here we introduce the loss functions which we will use to evaluate different VaR estimation models. These loss functions can be divided into two categories, firm losses and regulatory losses. The former penalizes both covered and non-covered losses, but the latter penalizes just the non-covered losses. Regulators are only concerned about the number and the size of non-covered losses, but firms want to maximize their profit, and because of their opportunity cost of capital, it is not optimal for them to overestimate VaR.

\cite{lopez_1998} proposes a general form of the loss function for VaR estimation models
\begin{equation}
L_{\mathrm{t}}=\left\{\begin{array}{ll}
f\left(r_{t}, \mathrm{VaR_{t}}\right) & \text { if } r_{t}<\mathrm{VaR_{t}}, \\
g\left(r_{t}, \mathrm{VaR_{t}}\right) & \text { if } r_{i} \geq \mathrm{VaR_{t}},
\end{array}\right.
\end{equation} 
where $f\left(r_{t}, \mathrm{VaR}\right)$ is a function for cases when the real returns fall below the VaR estimates, and $g\left(r_{t}, \mathrm{VaR}\right)$ is a function for cases when the real returns are above the VaR estimate. Lopez then introduces the following loss function with a quadratic term for regulatory purposes
\begin{equation}
RQL_{t}=\left\{\begin{array}{cc}
1+\left(\mathrm{VaR_{t}}-r_{t}\right)^{2} & \text { if } r_{t}<\mathrm{VaR_{t}}, \\
0 \quad & \text { if } r_{t} \geq \mathrm{VaR_{t}}.
\end{array}\right.
\end{equation}
Linear and Quadratic loss functions are also two of the simplest loss functions for backtesting risk measurement models. The quadratic form penalizes larger deviations from the realized return more than the linear case. These two functions can be used as $f$ and $g$ in Lopez's generalized form as
\begin{equation}
\begin{aligned}
LL_t &= |r_{t}-\mathrm{VaR_{t}}|, \\
QL_t &=(r_{t}-\mathrm{VaR_{t}})^2.
\end{aligned}
\end{equation}
\cite{sarma_thomas_shah_2003} suggests the following loss function, in which a firm’s opportunity cost can be used to penalize covered losses
\begin{equation}
FS_{t}=\left\{\begin{array}{cc}
(\mathrm{VaR_{t}}-r_{t})^2 & \text { if }  r_{t} < \mathrm{VaR_{t}}, \\
-\beta \cdot \mathrm{VaR_{t}} & \text { if }   r_{t} \geq \mathrm{VaR_{t}}.
\end{array}\right.
\end{equation}
\cite{angelidis_benos_degiannakis_2004} defines the p-quantile of out-of-sample observations as a proxy for the "true" VaR at the $1-p$\% confidence level. This quantile loss has the form 
\begin{equation}
QL_{t}=\left
\{\begin{array}{ll}
\left(r_{t}-\mathrm{VaR_{t}}\right)^{2} & \text { if }  r_{t}<\mathrm{VaR_{t}}, \\
{\left[\text { Percentile }(r, 100 p)_{1}^{T}-\mathrm{VaR_{t}}\right]^{2}} & \text { if }  r_{t} \geq \mathrm{VaR_{t}}.
\end{array}
\right.
\end{equation}
\cite{caporin_2008} also defines three loss functions for VaR estimates
\begin{equation}
\begin{aligned}
^1CL_t &= |1-|\frac{r_{t}}{\mathrm{VaR_{t}}}|| \\
^2CL_t &=\frac{\left(\left|r_{t}\right|-\left|\mathrm{VaR_{t}}\right|\right)^{2}}{\left|\mathrm{VaR_{t}}\right|} \\
^3CL_t &=\left|r_{t}-\mathrm{VaR_{t}}\right|
\end{aligned}
\end{equation}
These functions can be used for both covered and non-covered losses. The first and third functions consider the ratio and difference between the return and risk estimates. The second function calculates the squared error of the VaR estimate and standardizes it with the VaR estimate. \cite{caporin_2008} claims that this function can be thought as a first and second order loss. In Table \ref{tab:loss_table_5}, only the results from the second loss function are presented. 

\cite{Sener_baronyan_menguturk_2012} propose a ranking model based on a loss function that is only focused on the negative return space. This space is divided into the safe and violation spaces. In this loss function, covered losses are only penalized when the return is negative (safe space). In violation space, not only the magnitude of unexpected losses, but also their clusters are evaluated and combined to produce a quantity called penalization measure. In this context, a cluster is defined as a sequence of unexpected losses. Despite Christopherson that tests the violation bunching by his well-known statistics, this kind of loss function penalizes the autocorrelation of the unexpected losses. A detailed formulation of this loss function is presented in the appendix.


\begin{figure}
\centering
\includegraphics[width=9.2cm]{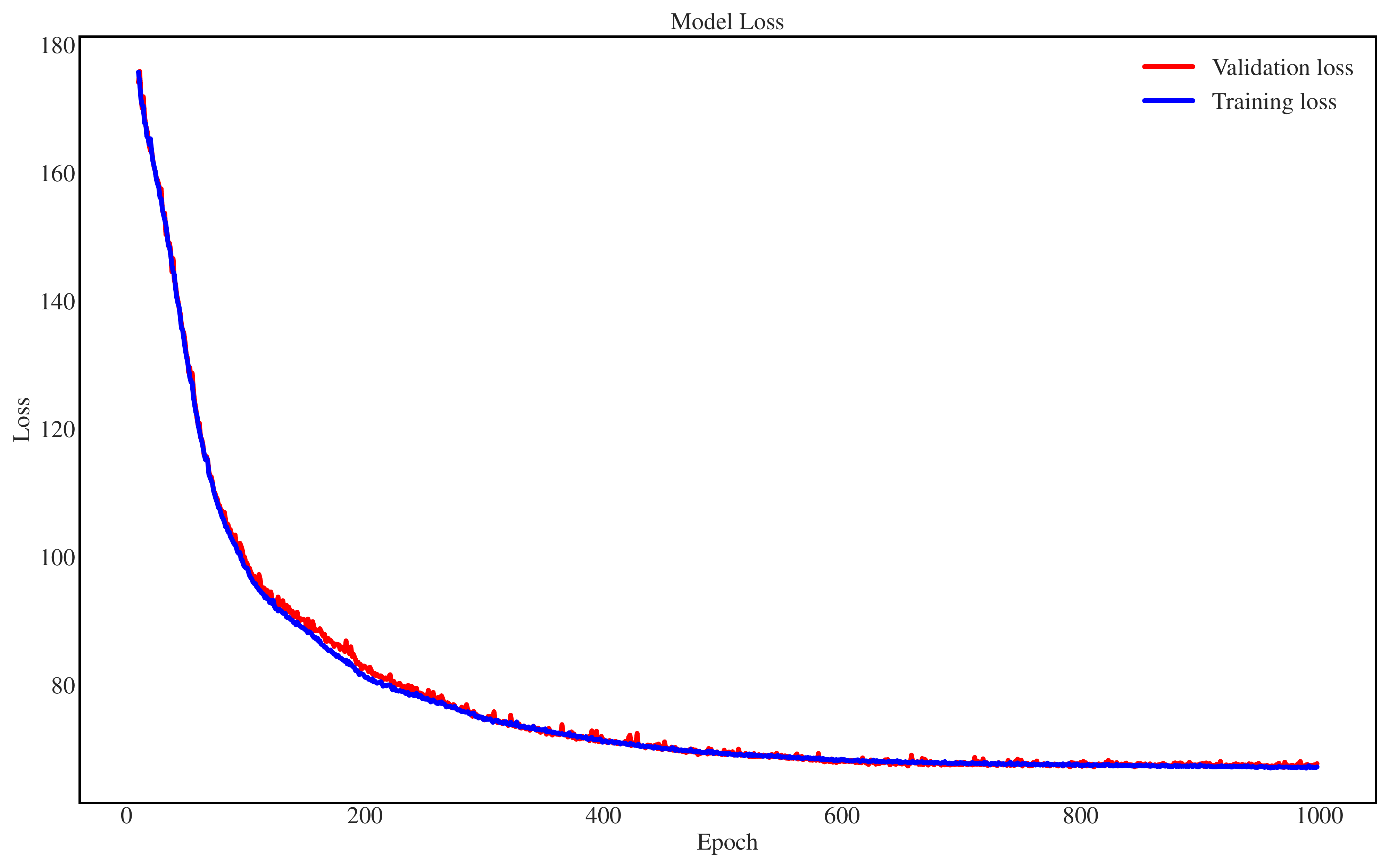}
\caption{VAE's loss for training and validation datasets}%
\label{fig:ann}%
\end{figure}

\subsection{Data}

To test our model on real data, we use portfolios from three different markets.  The data from these markets are extracted from the timeframe available in Alpha Vantage\footnote{\url{https://www.alphavantage.co}} and Quandl\footnote{\url{https://www.quandl.com}} websites. The stocks from these markets were chosen in a way that not more than $10\%$ of their data were missing.  This dataset includes a portfolio of 346 stocks included in S\&P500 index from December 11 1998, to March 27 2018 (4853 samples), a portfolio of 191 stocks from London Stock Exchange(LSE) from July 27 2000, to December 5 2019 (4754 samples), and a portfolio of 82 stocks from Frankfurt Stock Exchange(FSE) from January 2 2001, to October 29 2019.  For all of these portfolios, we use the first 75\% of the sample data for VAE training and the next 12.5\% for validation data. The last 12.5\% are used for testing the results of our model and comparing them with other models. 

The descriptive statistics of train and test data in all three portfolios are provided in Table \ref{table:statistics}. The kurtoses of these portfolios imply that their distributions have heavier tails than normal distribution. The Jarque-Bera statistics are also significant, which means that the normality assumption is rejected. 

\begin{table}
  \centering
  \caption{Descriptive statistics of the daily returns }
  \resizebox{\textwidth}{!}{%
    \begin{tabular}{lcccccccc}
    \toprule
          &       & \multicolumn{1}{l}{\textbf{Minimum}} & \multicolumn{1}{l}{\textbf{Maximum}} & \multicolumn{1}{l}{\textbf{Mean}} & \multicolumn{1}{l}{\textbf{ Std. Dev}} & \multicolumn{1}{l}{\textbf{Skewness}} & \multicolumn{1}{l}{\textbf{Kurtosis}} & \textbf{Jarque-Bera(p-value)} \\
    \midrule
    \multirow{2}[0]{*}{\textbf{S\&P500}} & train & -0.1024 & 0.112145 & 0.00066992 & 0.01357 & 0.004397 & 11.67114586 & 11353.51(0.00) \\
          & test  & -0.0378 & 0.027418 & 0.00046242 & 0.00773 & -0.56919 & 6.04903948 & 266.57(0.00) \\
    \midrule
    \multirow{2}[0]{*}{\textbf{LSE}} & train & -0.0593 & 0.1290159 & 0.00054038 & 0.01002 & 0.707229 & 18.05877001 & 33828.95(0.00) \\
          & test  & -0.0261 & 0.0341889 & -0.00002161 & 0.00621 & -0.17442 & 5.89436064 & 209.64(0.00) \\
    \midrule
    \multirow{2}[0]{*}{\textbf{FSE}} & train & -0.0803 & 0.122574 & 0.0004649 & 0.01342 & 0.162159 & 9.87235893 & 7017.33(0.00) \\
          & test  & -0.034 & 0.0332647 & -0.00027967 & 0.00914 & -0.30061 & 3.69386475 & 20.86(2.94e-5) \\
    \bottomrule
    \end{tabular}%
    }
    \label{table:statistics}
\end{table}%

\subsection{Training the model}
As stated in the methodology section, VAE is trained using the standardized data. Figure \ref{fig:ann} shows how the learning process has decreased the model's error. 

\begin{figure}[tbh]
\centering
\subfigure[Before training]{{\includegraphics[width=10.2cm]{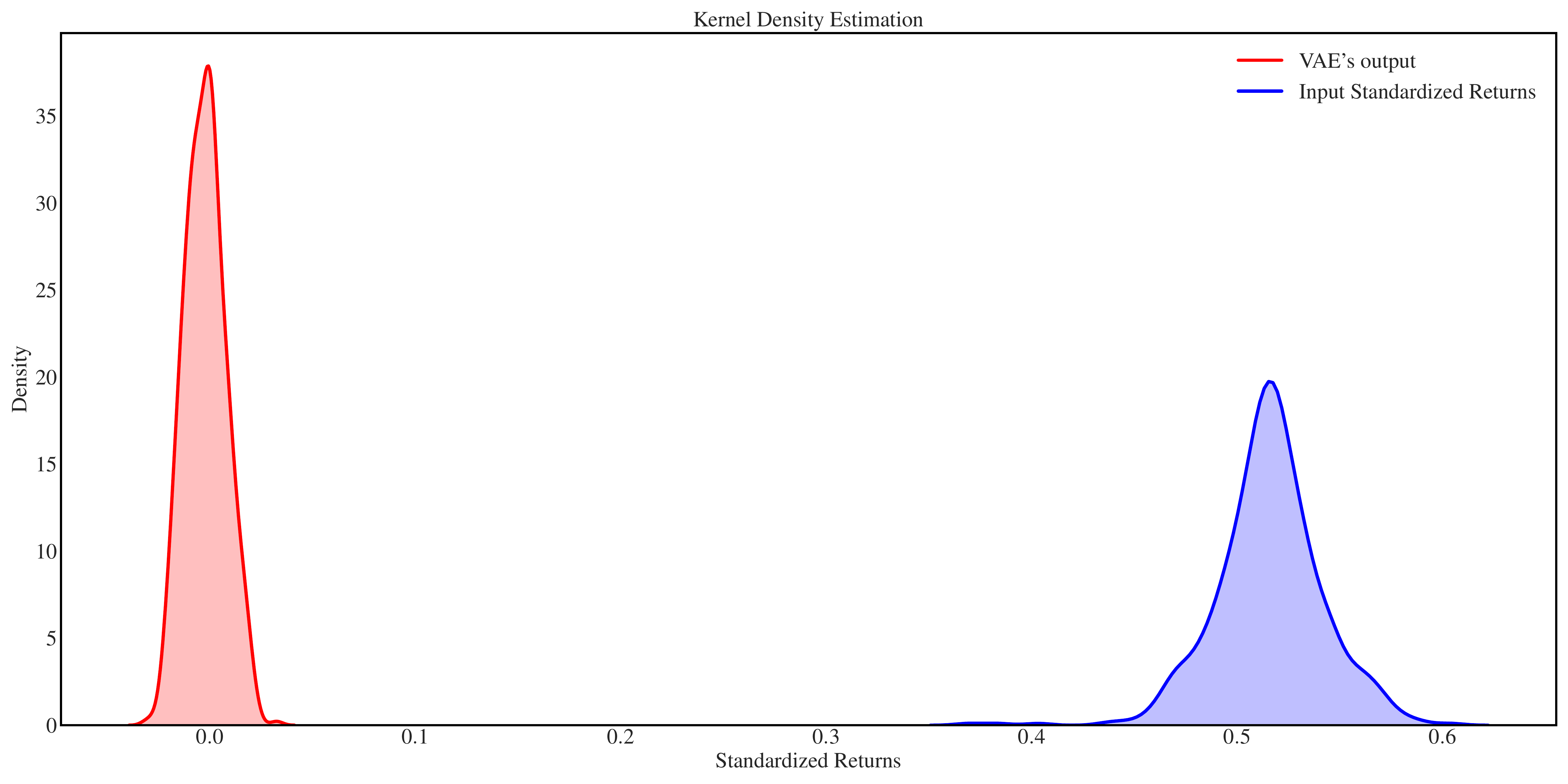}}}
\subfigure[After training]{{\includegraphics[width=10.2cm]{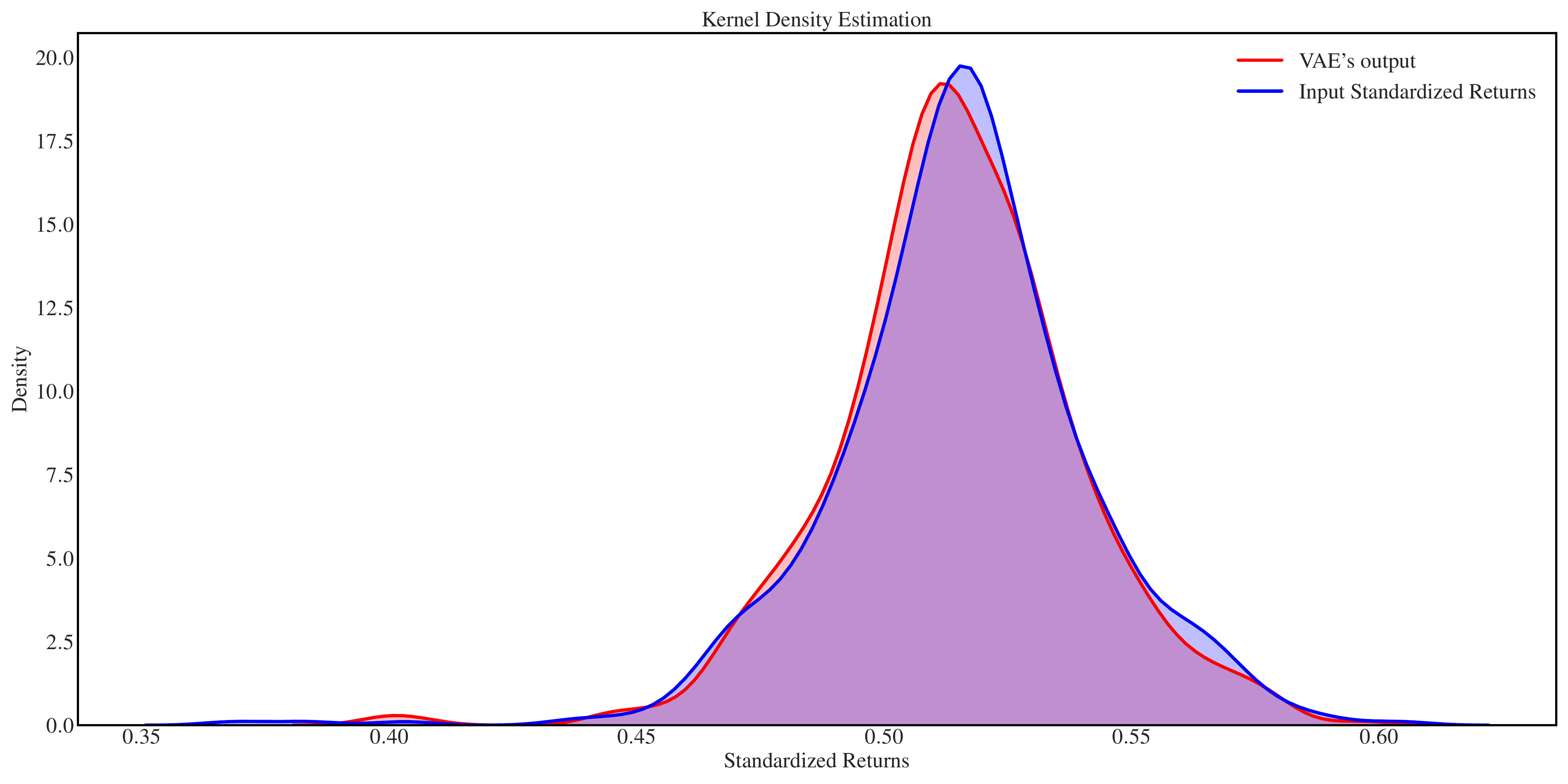}}}
\caption{Kernel density estimations for actual standardized returns vs. VAE's reconstructed returns (testing period, S\&P500 portfolio)\protect\footnotemark}
\label{fig:KDE}
\end{figure}

In figure \ref{fig:KDE}, we can see that the model learns to reconstruct the input returns very well. The larger the latent space, the better the model's reconstruction ability. However, it is also more computationally expensive to generate new stock returns when the size of the latent space increases. Furthermore, by using the Henze-Zirkler multivariate normality test (\cite{henze1990class}), we can see that the latent space has a mulivariate normal distribution which makes sampling possible. 

\footnotetext{Features are scaled once again to be in same range of [0,1] using Min-Max normalization}

\subsection{Correlation Matrix Analysis}
In this section, we first analyze the correlation matrix of real and generated returns and their signal-to-noise ratio. Then, we investigate how much the model is able to preserve the signal in the real returns.

\begin{figure}
\centering
\includegraphics[width=9.2cm]{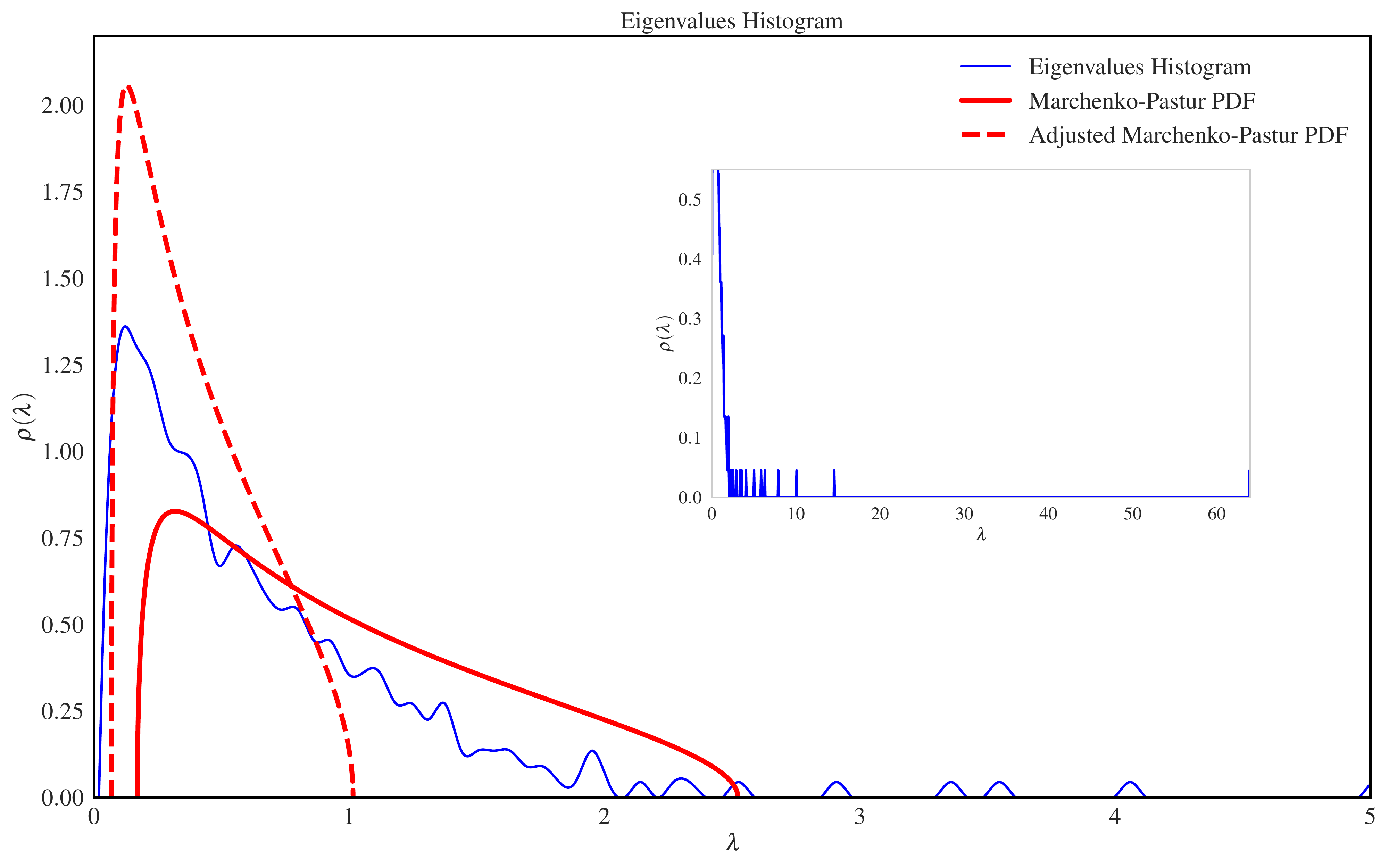}
\caption{Histogram of eigenvalues and Marchenko-Pastur PDF for real data (S\&P500 portfolio)}%
\label{fig:mpreal}%
\end{figure}

Suppose we have a matrix of independent and identically distributed random observations $X$, of size $T\times N$, with zero mean and variance $\sigma^2$, along with its covariance matrix $C$\footnote{When $\sigma^2 = 1$, $C$ becomes the correlation matrix of $X$.} \cite{marcenko_pastur_1967} proved that the eigenvalues of $C$ has the following probability density function, when $T,N \rightarrow \infty$ and $1< \frac{T}{N}<+\infty$, 

\begin{equation}
f[\lambda]=\left\{\begin{array}{ll}
\frac{T}{N} \frac{\sqrt{\left(\lambda_{+}-\lambda\right)\left(\lambda-\lambda_{-}\right)}}{2 \pi \lambda \sigma^{2}} & \text { if } \lambda \in\left[\lambda_{-}, \lambda_{+}\right] \\
0 & \text { if } \lambda \notin\left[\lambda_{-}, \lambda_{+}\right]
\end{array}\right.\end{equation}
where $\lambda_{\pm}=\sigma^{2}(1 \pm \sqrt{N / T})^{2} $. 
However, it can be seen in figure \ref{fig:mpreal} that the biggest eigenvalue of empirical correlation matrix\footnote{for S\&P500 portfolio} which can be associated to Market is almost 25 times bigger than the predicted $\lambda_{+}$. \cite{laloux_cizeau_potters_bouchaud_2000} suggest to treat $\sigma^2$ as an adjustable parameter to subtract the contribution of the eigenvalues above $\lambda_{+}$ that might contain some information. We use the algorithm suggested in \cite{prado_2020} to find the best fitted distribution, which implies that $\sigma^2 \approx 0.63$. This shows that only $37\%$ of the variance can be attributed to signal. 

\begin{figure}
\centering
\includegraphics[width=9.2cm]{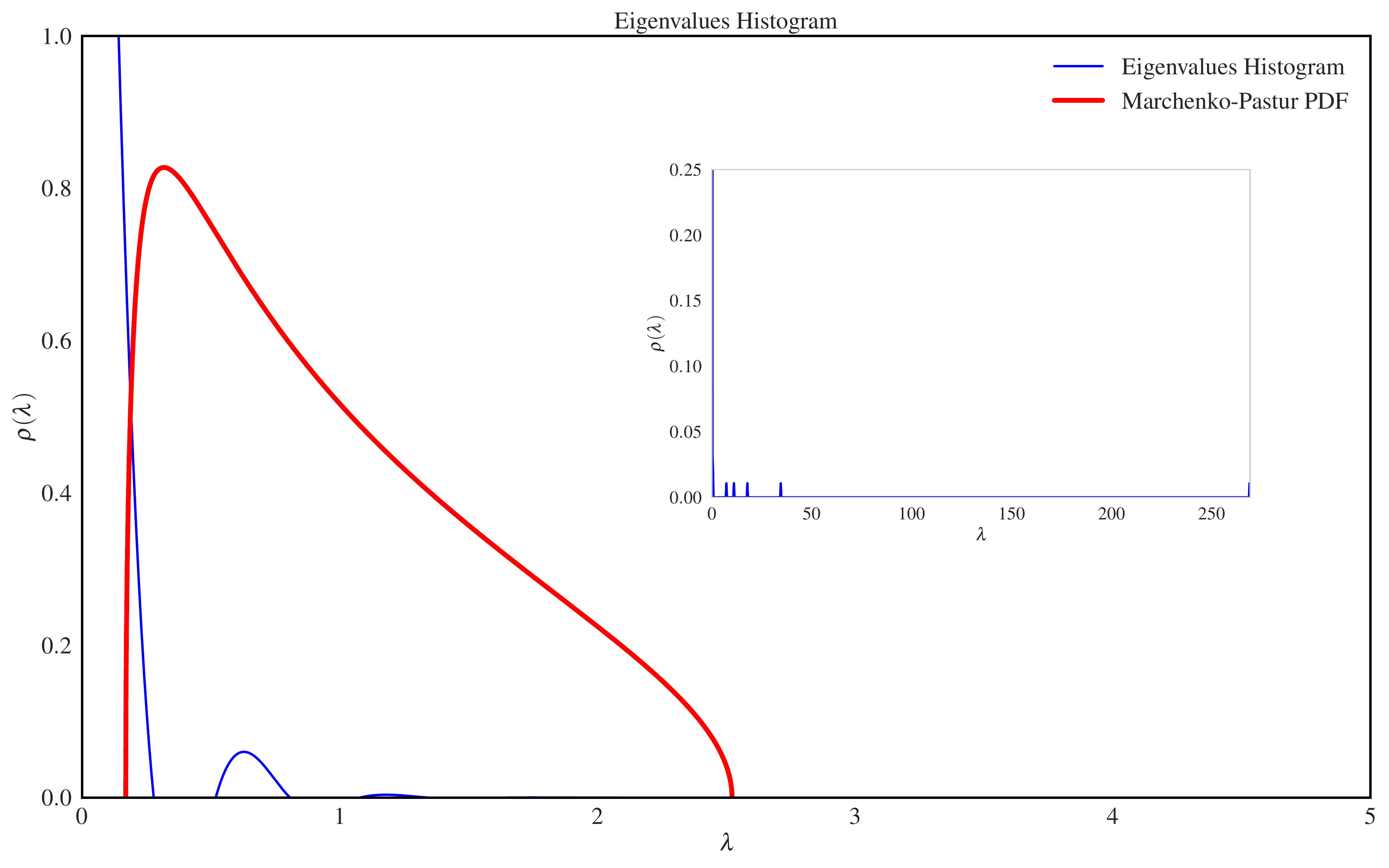}
\caption{Histogram of eigenvalues and Marchenko-Pastur PDF for generated data (S\&P500 portfolio)}%
\label{fig:mpgen}%
\end{figure}

Repeating the above-mentioned algorithm for generated returns shows that $98\%$ of the variance is explained by its signal eigenvectors and therefore Encoded VaR's output contains not much noise (Figure \ref{fig:mpgen}).

\begin{figure}
\centering
\includegraphics[width=9.2cm]{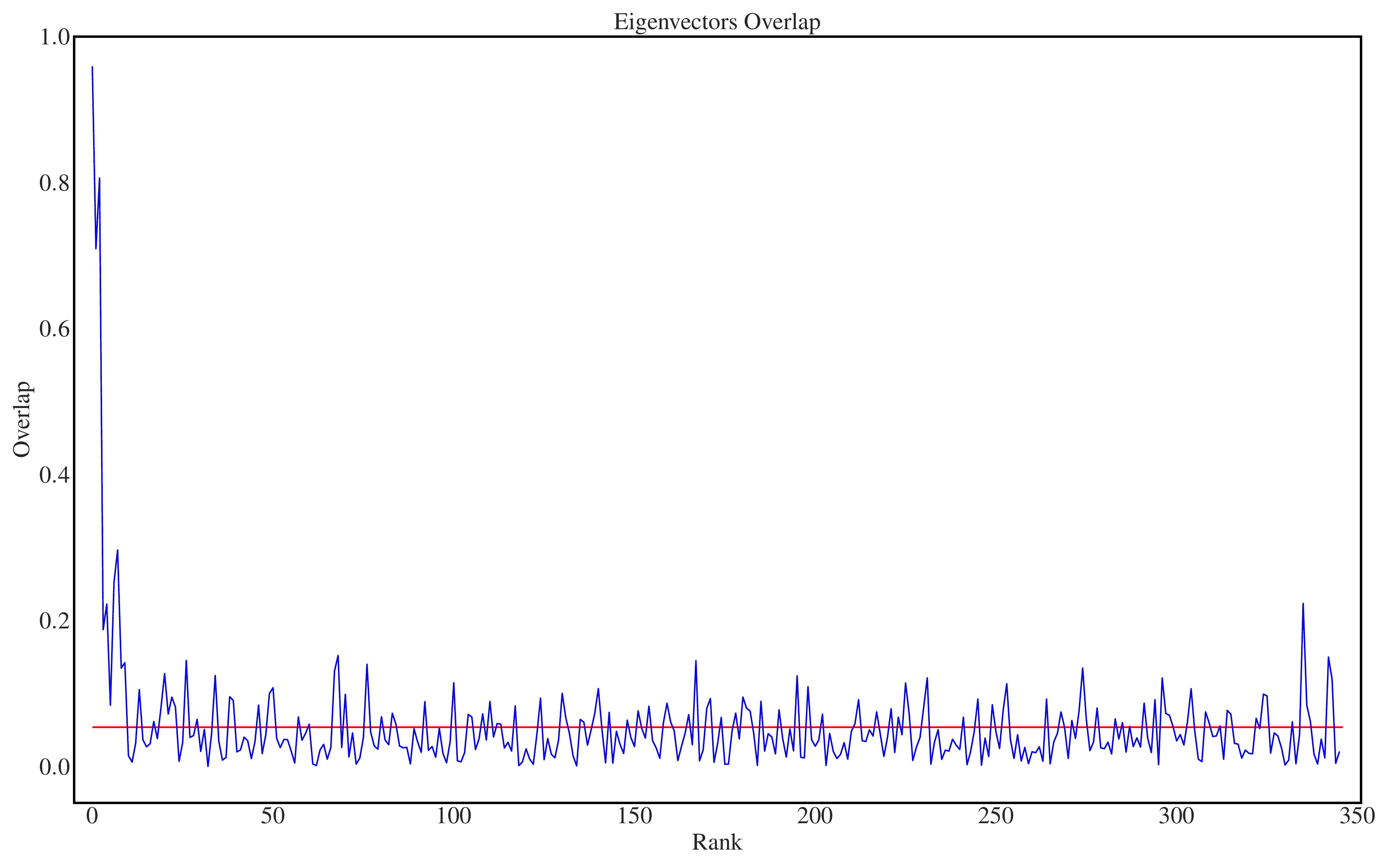}
\caption{Eigenvector overlap between real data and generated data(S\&P500 dataset), as a function of their rank n. The first 10 eigenvectors of real and generated data have dot product of 0.95, 0.71, 0.80, 0.19, 0.22, 0.08,
 0.25, 0.29, 0.13, and 0.14. The red horizontal line is the $1/\sqrt{N}$ threshold (0.05 for S\&P500 dataset)}%
\label{fig:overlap}%
\end{figure}

However, a better way to assess the generated returns is to calculate the amount of variance explained by signal eigenvectors of real data (and not generated data). We do this by measuring the variance of generated returns projected onto the signal eigenvectors, which shows that $79\%$ of the generated return's variance can be explained by real data's signal eigenvectors. Another way to compare these two correlation matrices is to calculate the eigenvector overlap between them, as a function of their rank $n$. This is done by calculating the scalar product of $n$th eigenvectors of both correlation matrices. As shown in Figure \ref{fig:overlap}, the first $10$ eigenvectors clearly have high overlap over the $1/\sqrt{N}$ threshold proposed by \cite{laloux_cizeau_potters_bouchaud_2000}.  These results show that not only the Encoded VaR model is able to learn the dependency structure of returns, but also it contains much less noise compared to real data. 

\subsection{Results}
The results of evaluating different loss functions for VaR models at  95\% and 99\% confidence levels are presented in Tables 2 and 3. From these results, it is clear that based on different loss functions, no model is superior to others in all portfolios. Different loss functions represent different points of view and with each one, we get a different ranking. However, there are some patterns that can be seen in these results. 

Despite its simplicity, Historical VaR is one of the top three models in terms of regulatory loss functions (except for 99\% VaR of S\&P500 portfolio), but it is not doing the same in terms of firm loss functions, which shows that Historical VaR  overestimates sometimes the risks of portfolio. Encoded VaR is also one of the top 5 models in terms of regulatory losses in all three portfolios, but unlike historical VaR, it performs well based on the firm loss functions, too. 

Since overestimation is not penalized by regulatory loss functions, we focus on the firm's loss functions to take into account this deficit. For 95\% VaR, we can see that Encoded VaR has outperformed most of the other models for FSE and LSE portfolios and it has one of the lowest losses among them. Even in S\&P500 portfolio for which Encoded VaR is ranked 6th in Sener's loss function, we can see that its loss is close to GARCH and E-GARCH models which are considered two of the best models for evaluating VaR.  

With 99\% confidence level, we can see once again that historical VaR has a very low loss in regulatory loss functions, especially in FSE and LSE portfolios. This is however, the result of risk overestimation, because this model has one of the highest Sener and Quantile losses. In these two portfolios, we can see that Encoded VaR has not only one of the lowest regulatory losses, but also it is performing well with respect to firm loss functions. For S\&P500 portfolio, Encoded VaR is also among the top 6 models in all loss functions and like the 95\% confidence level, its result are close to CaViaR and GARCH models.

Overall, it can be seen that Encoded VaR has one of the lowest losses compared to the benchmark models, and its results are close to some of the best models developed for VaR estimation like E-GARCH and CaViaR. 

\begin{table}
\centering
\caption{1-day 95\% VaR results with their comparative ranking for S\&P500, LSE and FSE}
\resizebox{\textwidth}{!}{%
\begin{tabular}{lccccccccc}
\toprule
& \multicolumn{8}{c}{\textbf{S\&P500}} \\
\cmidrule{2-9} & \textbf{Lopez} & \textbf{Linear} & \textbf{Quadratic} & \textbf{Caporin(R)} & \textbf{Sener} & \textbf{Sarma} & \textbf{Quantile } & \textbf{Caporin(F)} \\
\midrule
\textbf{RiskMetrics} & 0.0497(11) & 0.1653(10) & 0.0024(10) & 0.2252(10) & 0.0608(8) & 0.0607(11) & 0.0255(9) & 281.319(9) \\
    \textbf{Variance - Covariance} & 0.0447(6) & 0.186(11) & 0.003(11) & 0.3545(11) & 0.069(10) & 0.0559(7) & 0.0282(11) & 288.2136(11) \\
    \textbf{Historical} & 0.0348(1) & 0.1275(2) & 0.0016(2) & 0.1048(1) & 0.0752(11) & 0.0503(2) & 0.0165(2) & 198.9427(1) \\
    \textbf{Filtered Historical} & 0.0464(8) & 0.1503(7) & 0.002(7) & 0.2124(9) & 0.0563(3) & 0.0578(9) & 0.0331(12) & 287.0068(10) \\
    \textbf{Monte Carlo} & 0.0447(6) & 0.1877(12) & 0.003(12) & 0.3629(12) & 0.0689(9) & 0.0558(6) & 0.0279(10) & 289.8769(12) \\
    \textbf{GARCH} & 0.0464(8) & 0.1574(8) & 0.002(6) & 0.1674(7) & 0.0567(5) & 0.0574(8) & 0.0197(6) & 270.6803(7) \\
    \textbf{E-GARCH} & 0.0546(12) & 0.162(9) & 0.0021(9) & 0.1841(8) & 0.06(7) & 0.0654(12) & 0.019(5) & 275.4838(8) \\
    \textbf{CAViaR Symmetric} & 0.043(5) & 0.1437(5) & 0.0019(5) & 0.1473(4) & 0.0563(3) & 0.0549(5) & 0.0172(3) & 248.934(5) \\
    \textbf{CAViaR Asymmetric} & 0.048(10) & 0.1314(3) & 0.0016(1) & 0.153(5) & 0.0558(2) & 0.0598(10) & 0.0237(8) & 256.8689(6) \\
    \textbf{CAViaR Garch} & 0.0364(2) & 0.133(4) & 0.0017(3) & 0.1291(2) & 0.0547(1) & 0.0487(1) & 0.0178(4) & 241.7056(4) \\
    \textbf{CAViaR Adaptive} & 0.0381(3) & 0.1463(6) & 0.002(8) & 0.1565(6) & 0.0764(12) & 0.0528(4) & 0.0226(7) & 210.9561(2) \\
    \textbf{Encoded VaR} & 0.0389(4) & 0.1176(1) & 0.0017(4) & 0.1459(3) & 0.0568(6) & 0.0516(3) & 0.0163(1) & 231.1832(3) \\
\midrule
& \multicolumn{8}{c}{\textbf{LSE}} \\
\cmidrule{2-9} & \textbf{Lopez} & \textbf{Linear} & \textbf{Quadratic} & \textbf{Caporin(R)} & \textbf{Sener} & \textbf{Sarma} & \textbf{Quantile } & \textbf{Caporin(F)} \\
\midrule
\textbf{RiskMetrics} & 0.0676(11) & 0.1654(11) & 0.0014(10) & 0.1687(11) & 0.0735(10) & 0.0766(11) & 0.0082(8) & 286.2216(11) \\
    \textbf{Variance - Covariance} & 0.0591(8) & 0.1595(9) & 0.0014(8) & 0.1626(9) & 0.0688(8) & 0.0683(8) & 0.0059(3) & 281.3514(9) \\
    \textbf{Historical} & 0.022(1) & 0.0695(2) & 0.0006(3) & 0.0438(3) & 0.0527(5) & 0.0353(1) & 0.0126(11) & 197.5838(1) \\
    \textbf{Filtered Historical} & 0.0946(12) & 0.2111(12) & 0.0016(12) & 0.2147(12) & 0.0938(12) & 0.1023(12) & 0.0112(9) & 347.0906(12) \\
    \textbf{Monte Carlo} & 0.0608(9) & 0.161(10) & 0.0014(9) & 0.1627(10) & 0.0691(9) & 0.0699(9) & 0.0059(2) & 282.521(10) \\
    \textbf{GARCH} & 0.0405(3) & 0.0979(6) & 0.0008(6) & 0.0762(6) & 0.0532(6) & 0.0515(3) & 0.0068(5) & 236.3934(3) \\
    \textbf{E-GARCH} & 0.0473(7) & 0.123(7) & 0.0011(7) & 0.1077(7) & 0.0609(7) & 0.0574(7) & 0.0073(6) & 254.3721(7) \\
    \textbf{CAViaR Symmetric} & 0.0456(6) & 0.0878(4) & 0.0007(4) & 0.0636(4) & 0.0511(4) & 0.0564(6) & 0.0119(10) & 241.1616(4) \\
    \textbf{CAViaR Asymmetric} & 0.0422(4) & 0.0672(1) & 0.0004(1) & 0.0414(1) & 0.0418(2) & 0.0528(5) & 0.0137(12) & 250.1357(6) \\
    \textbf{CAViaR Garch} & 0.0422(5) & 0.0965(5) & 0.0007(5) & 0.0673(5) & 0.0477(3) & 0.0527(4) & 0.0081(7) & 245.4106(5) \\
    \textbf{CAViaR Adaptive} & 0.0608(9) & 0.1584(8) & 0.0015(11) & 0.1585(8) & 0.0744(11) & 0.0701(10) & 0.0024(1) & 276.9831(8) \\
    \textbf{Encoded VaR} & 0.0372(2) & 0.0742(3) & 0.0005(2) & 0.0431(2) & 0.0413(1) & 0.0479(2) & 0.0065(4) & 234.3024(2) \\
\midrule
& \multicolumn{8}{c}{\textbf{FSE}} \\
\cmidrule{2-9} & \textbf{Lopez} & \textbf{Linear} & \textbf{Quadratic} & \textbf{Caporin(R)} & \textbf{Sener} & \textbf{Sarma} & \textbf{Quantile } & \textbf{Caporin(F)} \\
\midrule
\textbf{RiskMetrics} & 0.0926(12) & 0.2421(11) & 0.0018(9) & 0.129(9) & 0.1093(11) & 0.1059(12) & 0.0209(10) & 281.7467(11) \\
    \textbf{Variance - Covariance} & 0.0774(6) & 0.2343(9) & 0.0019(10) & 0.1441(10) & 0.095(7) & 0.091(6) & 0.0193(7) & 281.4312(10) \\
    \textbf{Historical} & 0.0286(1) & 0.0929(2) & 0.0007(2) & 0.0493(1) & 0.0919(6) & 0.0499(1) & 0.0181(4) & 189.549(1) \\
    \textbf{Filtered Historical} & 0.0842(9) & 0.197(6) & 0.0013(6) & 0.1(6) & 0.0918(5) & 0.0983(9) & 0.0262(12) & 273.0914(7) \\
    \textbf{Monte Carlo} & 0.0791(7) & 0.24(10) & 0.0019(11) & 0.1484(11) & 0.0972(8) & 0.0926(7) & 0.0193(8) & 283.2139(12) \\
    \textbf{GARCH} & 0.0892(11) & 0.2267(8) & 0.0017(8) & 0.1207(8) & 0.0994(10) & 0.1025(11) & 0.0198(9) & 278.7519(9) \\
    \textbf{E-GARCH} & 0.0875(10) & 0.2214(7) & 0.0016(7) & 0.1179(7) & 0.0979(9) & 0.101(10) & 0.019(6) & 275.056(8) \\
    \textbf{CAViaR Symmetric} & 0.0606(3) & 0.1419(4) & 0.001(5) & 0.0656(5) & 0.0737(3) & 0.0764(3) & 0.017(3) & 240.0026(3) \\
    \textbf{CAViaR Asymmetric} & 0.0471(2) & 0.0839(1) & 0.0006(1) & 0.0493(2) & 0.0599(1) & 0.0634(2) & 0.0233(11) & 236.0036(2) \\
    \textbf{CAViaR Garch} & 0.0673(5) & 0.1469(5) & 0.001(4) & 0.0619(4) & 0.0763(4) & 0.0828(5) & 0.0189(5) & 245.4999(4) \\
    \textbf{CAViaR Adaptive} & 0.0808(8) & 0.2535(12) & 0.0022(12) & 0.1663(12) & 0.1239(12) & 0.0949(8) & 0.0153(1) & 272.1478(6) \\
    \textbf{Encoded VaR} & 0.064(4) & 0.1322(3) & 0.0009(3) & 0.0551(3) & 0.0636(2) & 0.0785(4) & 0.0165(2) & 255.8279(5) \\
\bottomrule
\end{tabular}
}
\label{tab:loss_table_5}%
\vspace{1ex}\\
{\justifying \noindent Note: This table presents back-testing results of various VaR methods, including RiskMetrics, Variance-Covariance, GARCH, Historical and Filtered Historical Simulation, Monte Carlo Simulation, GARCH, EGARCH, CaviaR symmetric, CaviaR asymmetric, CaviaR Garch, CaviaR Adaptive and Encoded VaR.  
\par}
\end{table}%

\begin{table}
\centering
\caption{1-day 99\% VaR results with their comparative ranking for S\&P500, LSE and FSE}
\resizebox{\textwidth}{!}{%
\begin{tabular}{lcccccccc}
\toprule
& \multicolumn{8}{c}{\textbf{S\&P500}} \\
\cmidrule{2-9} & \textbf{Lopez} & \textbf{Linear} & \textbf{Quadratic} & \textbf{Caporin(R)} & \textbf{Sener} & \textbf{Sarma} & \textbf{Quantile } & \textbf{Caporin(F)} \\

\midrule
\textbf{RiskMetrics} & 0.0149(8) & 0.0865(10) & 0.0013(10) & 0.0901(10) & 0.0169(5) & 0.0311(6) & 0.0962(11) & 193.9534(10) \\
    \textbf{Variance - Covariance} & 0.0215(11) & 0.1098(11) & 0.0018(11) & 0.1524(11) & 0.0264(12) & 0.038(12) & 0.0893(9) & 194.4368(11) \\
    \textbf{Historical} & 0.0099(5) & 0.0574(6) & 0.0008(6) & 0.0365(4) & 0.022(9) & 0.0333(10) & 0.0106(1) & 133.2436(2) \\
    \textbf{Filtered Historical} & 0.0116(7) & 0.0618(7) & 0.001(9) & 0.0687(9) & 0.0146(3) & 0.0302(3) & 0.1072(12) & 170.9601(7) \\
    \textbf{Monte Carlo} & 0.0215(11) & 0.1107(12) & 0.0018(12) & 0.1552(12) & 0.0263(11) & 0.0378(11) & 0.0894(10) & 196.3675(12) \\
    \textbf{GARCH} & 0.0166(9) & 0.0634(8) & 0.0009(7) & 0.0533(7) & 0.0145(2) & 0.0329(9) & 0.0772(7) & 182.4548(8) \\
    \textbf{E-GARCH} & 0.0166(9) & 0.0713(9) & 0.001(8) & 0.062(8) & 0.0153(4) & 0.0327(8) & 0.0782(8) & 185.943(9) \\
    \textbf{CAViaR Symmetric} & 0.0099(5) & 0.0484(5) & 0.0005(3) & 0.0251(3) & 0.0177(8) & 0.0303(5) & 0.0545(5) & 143.9189(5) \\
    \textbf{CAViaR Asymmetric} & 0.0066(2) & 0.0426(3) & 0.0007(4) & 0.0504(6) & 0.0139(1) & 0.0255(1) & 0.0663(6) & 157.2935(6) \\
    \textbf{CAViaR Garch} & 0.0099(4) & 0.0423(2) & 0.0004(2) & 0.0185(2) & 0.0172(7) & 0.0303(4) & 0.0508(4) & 142.5759(3) \\
    \textbf{CAViaR Adaptive} & 0.0066(1) & 0.0326(1) & 0.0003(1) & 0.0135(1) & 0.0225(10) & 0.032(7) & 0.0316(2) & 112.3972(1) \\
    \textbf{Encoded VaR} & 0.0066(2) & 0.0482(4) & 0.0008(5) & 0.044(5) & 0.017(6) & 0.0268(2) & 0.0461(3) & 142.9084(4) \\

\midrule
& \multicolumn{8}{c}{\textbf{LSE}} \\
\cmidrule{2-9} & \textbf{Lopez} & \textbf{Linear} & \textbf{Quadratic} & \textbf{Caporin(R)} & \textbf{Sener} & \textbf{Sarma} & \textbf{Quantile } & \textbf{Caporin(F)} \\

\midrule
\textbf{RiskMetrics} & 0.0287(10) & 0.0783(12) & 0.0006(10) & 0.0517(10) & 0.0187(9) & 0.042(11) & 0.03(7) & 8.1965(2) \\
    \textbf{Variance - Covariance} & 0.027(9) & 0.0752(10) & 0.0006(11) & 0.0529(11) & 0.0196(10) & 0.0404(9) & 0.0232(5) & 8.2298(3) \\
    \textbf{Historical} & 0.0017(1) & 0.0051(1) & 0.0000(1) & 0.0013(1) & 0.0208(12) & 0.0274(6) & 0.0471(12) & 15.2122(12) \\
    \textbf{Filtered Historical} & 0.0287(10) & 0.0646(9) & 0.0004(9) & 0.0315(9) & 0.0155(8) & 0.0424(12) & 0.0399(11) & 8.3857(4) \\
    \textbf{Monte Carlo} & 0.0287(10) & 0.0772(11) & 0.0007(12) & 0.0545(12) & 0.0202(11) & 0.0419(10) & 0.0235(6) & 8.1602(1) \\
    \textbf{GARCH} & 0.0101(6) & 0.0314(6) & 0.0003(7) & 0.0188(7) & 0.0121(2) & 0.0263(4) & 0.0149(2) & 9.7057(6) \\
    \textbf{E-GARCH} & 0.0152(8) & 0.0477(8) & 0.0004(8) & 0.028(8) & 0.0119(1) & 0.0302(8) & 0.0203(4) & 9.057(5) \\
    \textbf{CAViaR Symmetric} & 0.0084(5) & 0.0172(5) & 0.0001(4) & 0.0079(4) & 0.0132(5) & 0.0265(5) & 0.0331(8) & 10.7594(9) \\
    \textbf{CAViaR Asymmetric} & 0.0034(2) & 0.0135(3) & 0.0001(5) & 0.0089(5) & 0.0124(3) & 0.0205(1) & 0.0335(9) & 10.1933(7) \\
    \textbf{CAViaR Garch} & 0.0068(4) & 0.0142(4) & 0.0001(3) & 0.0059(3) & 0.0131(4) & 0.0247(3) & 0.0395(10) & 10.6803(8) \\
    \textbf{CAViaR Adaptive} & 0.0101(6) & 0.0343(7) & 0.0002(6) & 0.0148(6) & 0.015(7) & 0.0293(7) & 0.0112(1) & 11.501(11) \\
    \textbf{Encoded VaR} & 0.0055(3) & 0.0086(2) & 0.0000(2) & 0.002(2) & 0.0141(6) & 0.0241(2) & 0.0168(3) & 11.1123(10) \\

\midrule
& \multicolumn{8}{c}{\textbf{FSE}} \\
\cmidrule{2-9} & \textbf{Lopez} & \textbf{Linear} & \textbf{Quadratic} & \textbf{Caporin(R)} & \textbf{Sener} & \textbf{Sarma} & \textbf{Quantile } & \textbf{Caporin(F)} \\

\midrule
\textbf{RiskMetrics} & 0.0269(10) & 0.0531(10) & 0.0003(10) & 0.0156(10) & 0.0192(10) & 0.047(10) & 0.0336(9) & 193.1089(11) \\
    \textbf{Variance - Covariance} & 0.0286(11) & 0.0612(11) & 0.0004(11) & 0.0237(11) & 0.0188(8) & 0.0486(12) & 0.0299(7) & 193.0897(10) \\
    \textbf{Historical} & 0(1)  & 0(1)  & 0(1)  & 0(1)  & 0.027(12) & 0.0342(6) & 0.0792(12) & 104.8519(1) \\
    \textbf{Filtered Historical} & 0.0101(6) & 0.0188(6) & 0.0001(4) & 0.0057(5) & 0.0165(3) & 0.0335(5) & 0.0452(11) & 166.5123(7) \\
    \textbf{Monte Carlo} & 0.0286(11) & 0.0662(12) & 0.0004(12) & 0.0261(12) & 0.0189(9) & 0.0484(11) & 0.0304(8) & 195.9604(12) \\
    \textbf{GARCH} & 0.0236(8) & 0.0433(9) & 0.0003(9) & 0.0147(8) & 0.0158(2) & 0.0439(8) & 0.0298(6) & 188.0601(9) \\
    \textbf{E-GARCH} & 0.0236(8) & 0.0416(8) & 0.0003(8) & 0.015(9) & 0.0154(1) & 0.0442(9) & 0.0284(3) & 185.4893(8) \\
    \textbf{CAViaR Symmetric} & 0.0067(3) & 0.0161(3) & 0.0001(3) & 0.0052(4) & 0.0175(6) & 0.0313(2) & 0.0342(10) & 154.6601(5) \\
    \textbf{CAViaR Asymmetric} & 0.0067(3) & 0.0163(4) & 0.0001(6) & 0.0081(7) & 0.0176(7) & 0.0314(3) & 0.0294(4) & 151.276(3) \\
    \textbf{CAViaR Garch} & 0.0084(5) & 0.0178(5) & 0.0001(5) & 0.0051(3) & 0.0169(4) & 0.0323(4) & 0.0296(5) & 158.9313(6) \\
    \textbf{CAViaR Adaptive} & 0.0152(7) & 0.0261(7) & 0.0002(7) & 0.0069(6) & 0.021(11) & 0.0419(7) & 0.0227(1) & 144.0999(2) \\
    \textbf{Encoded VaR} & 0.0044(2) & 0.0071(2) & 0.0000(2) & 0.001(2) & 0.0171(5) & 0.0287(1) & 0.0271(2) & 153.1048(4) \\
\bottomrule
\end{tabular}
}
\label{tabDMNikkei_BVSP_Merval}%
\vspace{1ex}\\
{\justifying \noindent Note: This table presents back-testing results of various VaR methods, including RiskMetrics, Variance-Covariance, GARCH, Historical and Filtered Historical Simulation, Monte Carlo Simulation, GARCH, EGARCH, CaviaR symmetric, CaviaR asymmetric, CaviaR Garch, CaviaR Adaptive and Encoded VaR.  
     \par}
\end{table}%

\begin{figure}
    \centering
    \subfigure[CAViaR Garch]{{\includegraphics[width=7.7cm, height=2.1cm]{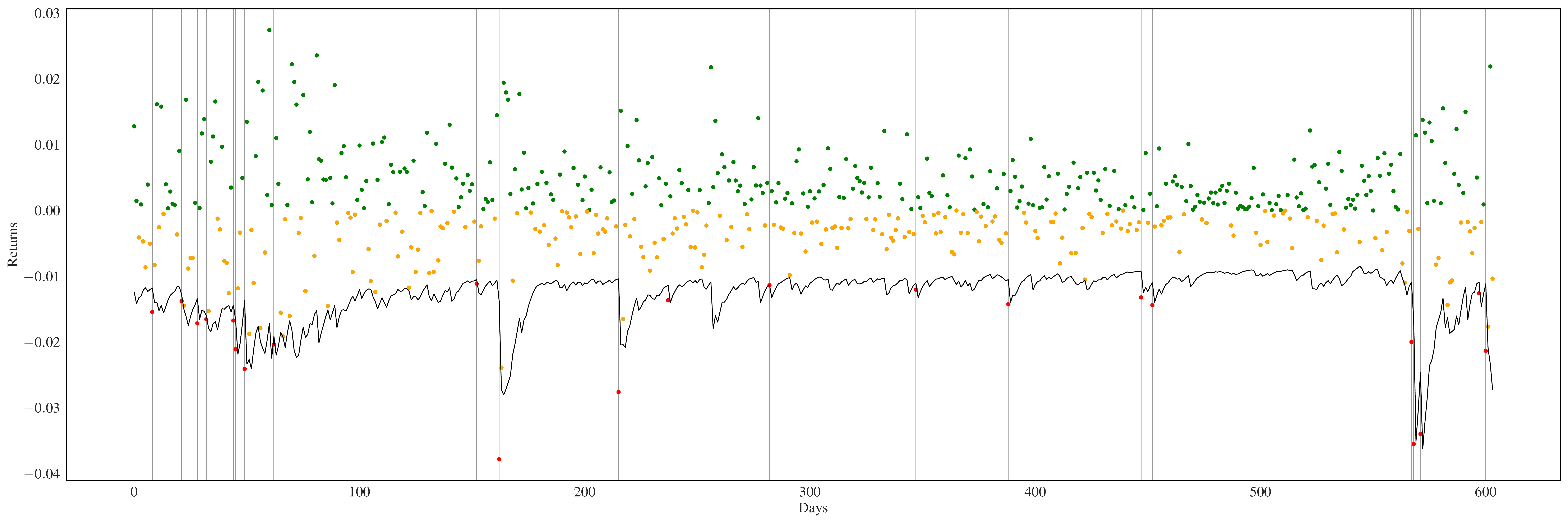}}}
    \subfigure[CAViaR Adaptive]{{\includegraphics[width=7.7cm, height=2.1cm]{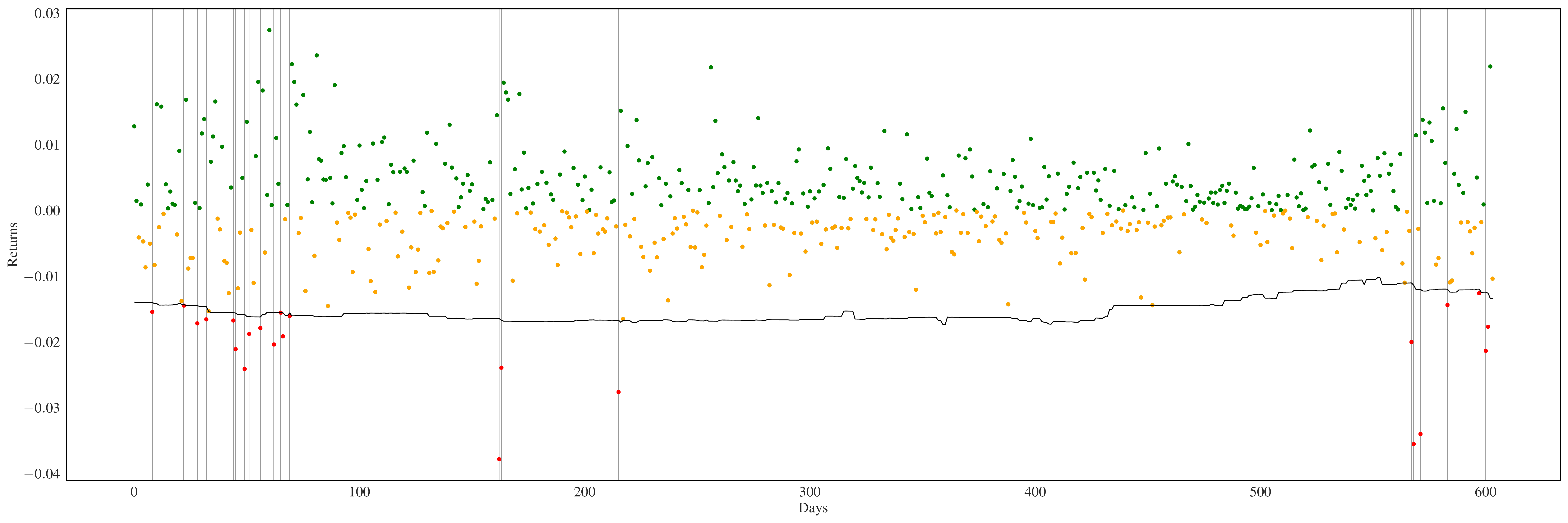}}}\\
    \subfigure[CAViaR Symmetric]{{\includegraphics[width=7.7cm, height=2.1cm]{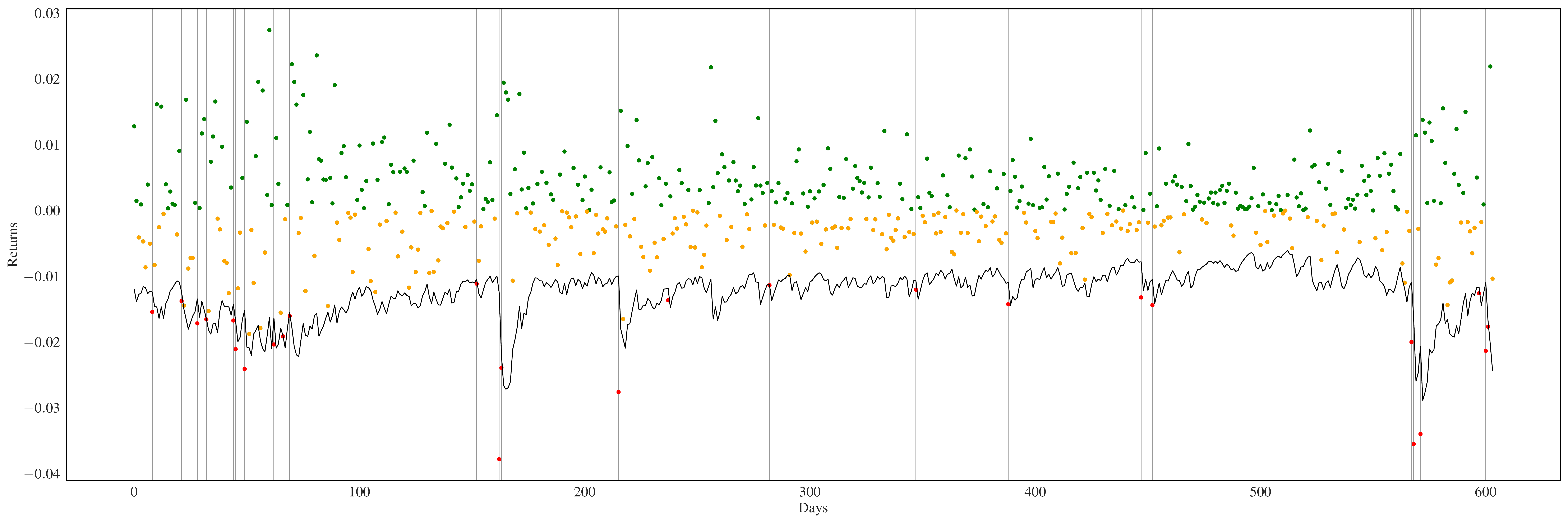}}}
    \subfigure[CAViaR Asymmetric]{{\includegraphics[width=7.7cm, height=2.1cm]{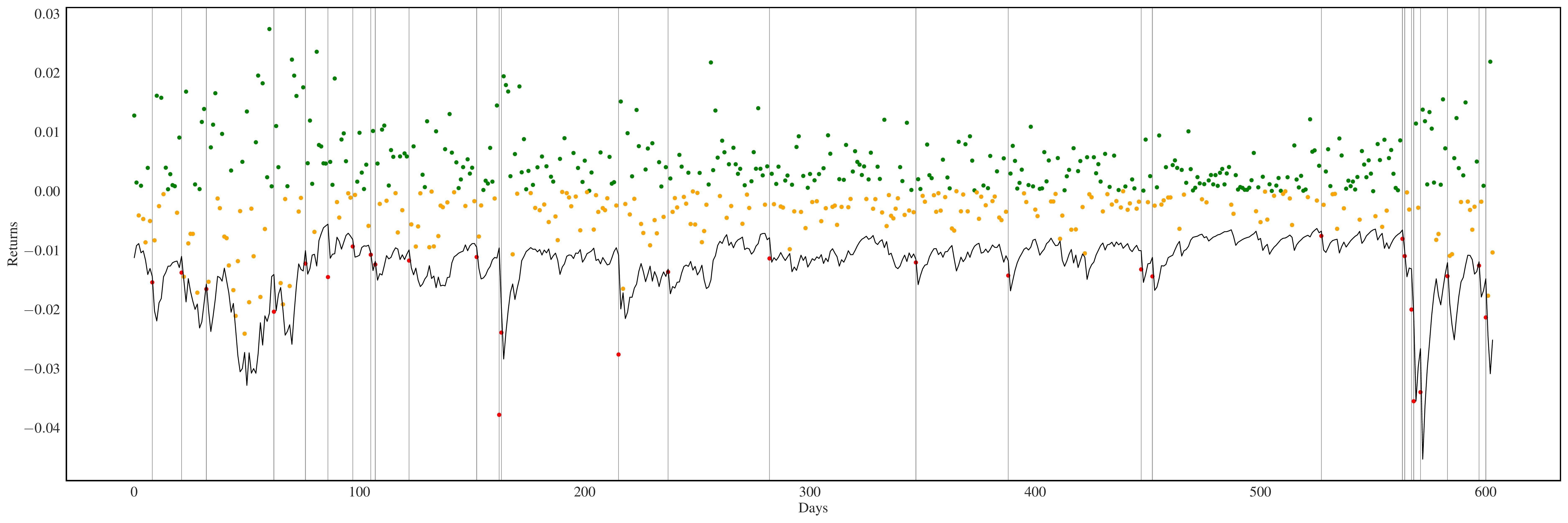}}}\\
    \subfigure[Historical]{{\includegraphics[width=7.7cm, height=2.1cm]{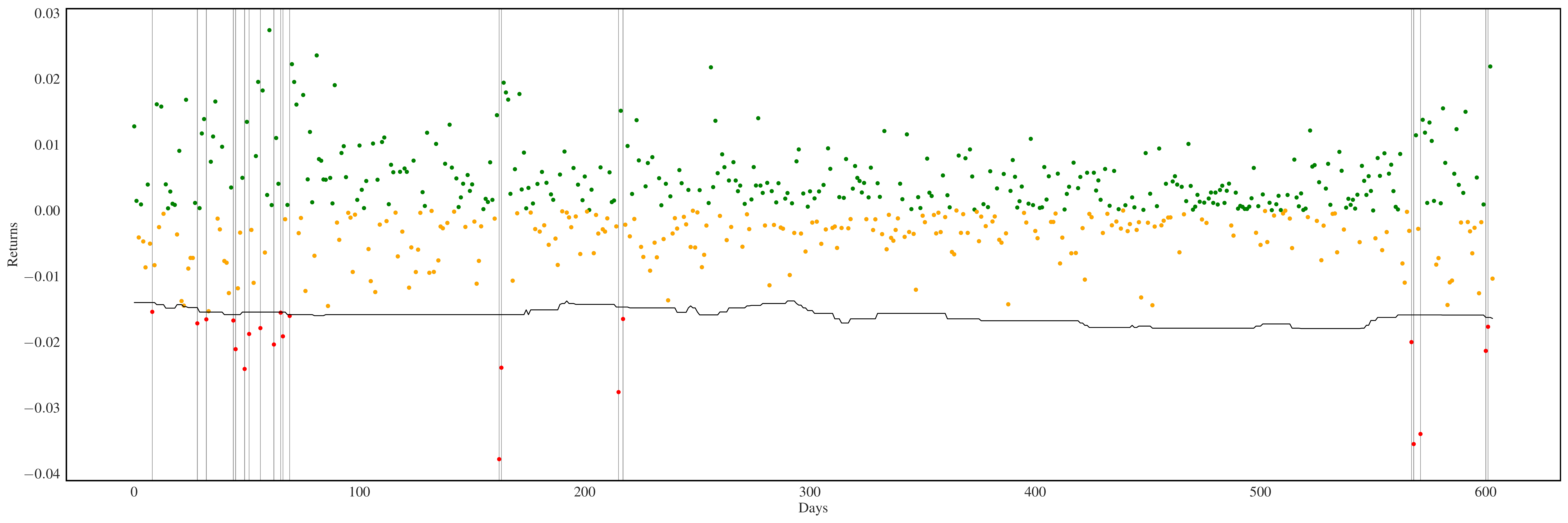}}}
    \subfigure[Filtered Historical]{{\includegraphics[width=7.7cm, height=2.1cm]{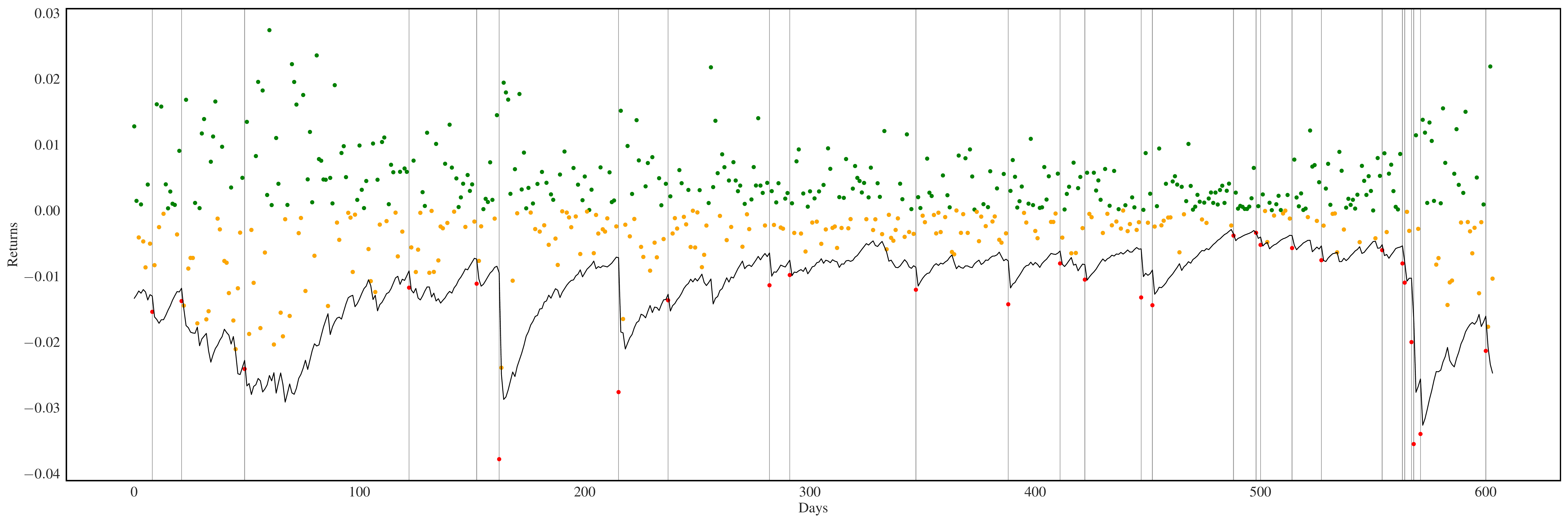}}}\\
    \subfigure[GARCH]{{\includegraphics[width=7.7cm, height=2.1cm]{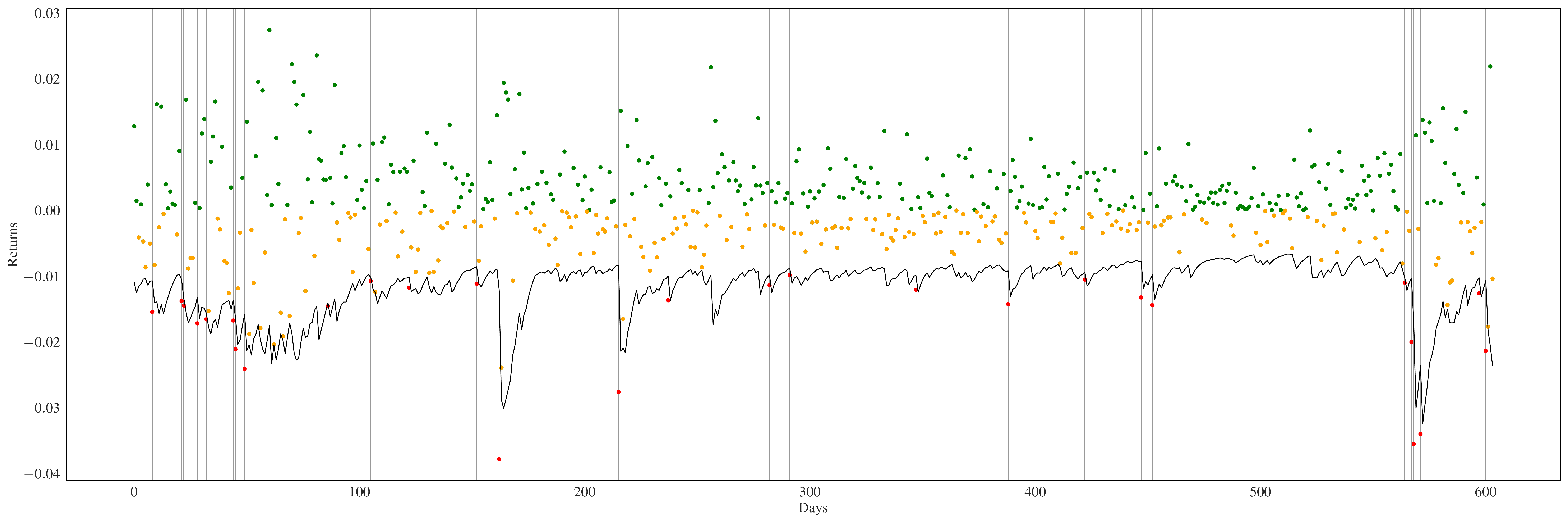}}}
    \subfigure[E-GARCH]{{\includegraphics[width=7.7cm, height=2.1cm]{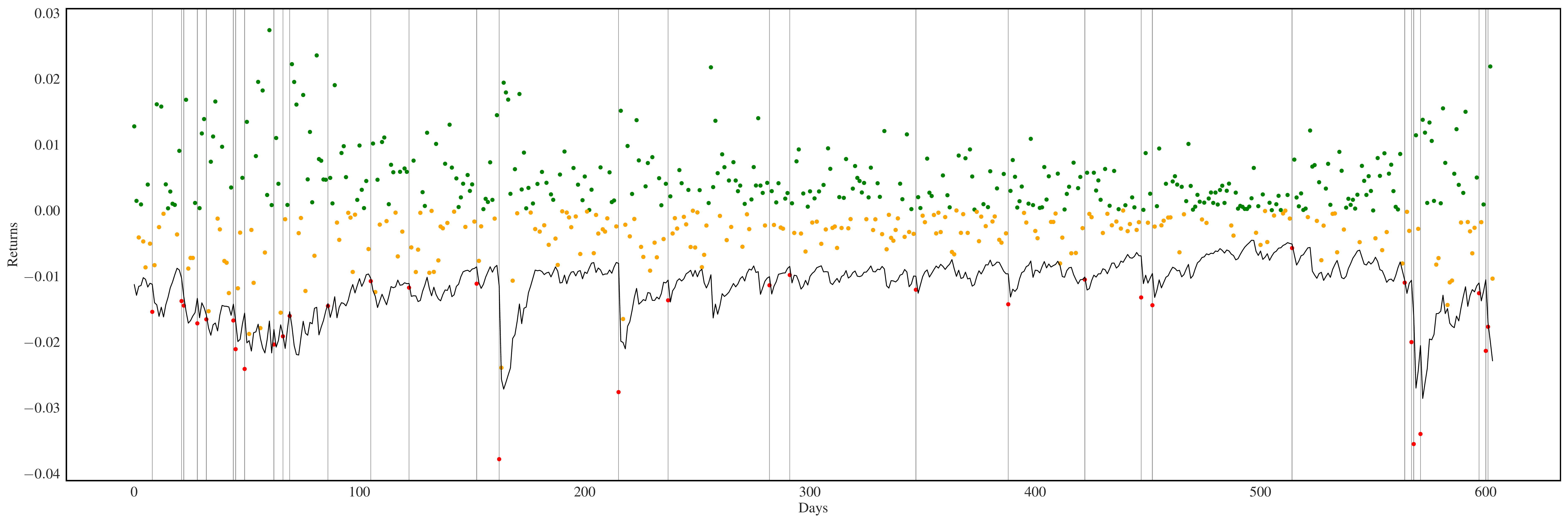}}}\\
    \subfigure[RiskMetrics]{{\includegraphics[width=7.7cm, height=2.1cm]{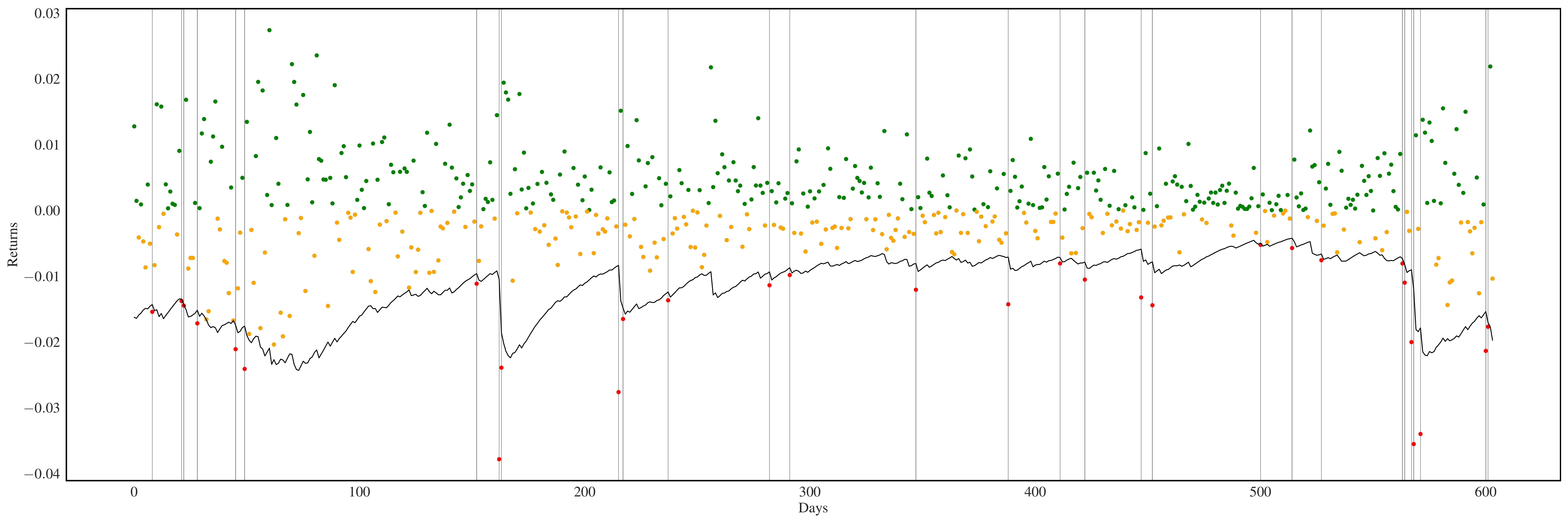}}}
    \subfigure[Variance Covariance]{{\includegraphics[width=7.7cm, height=2.1cm]{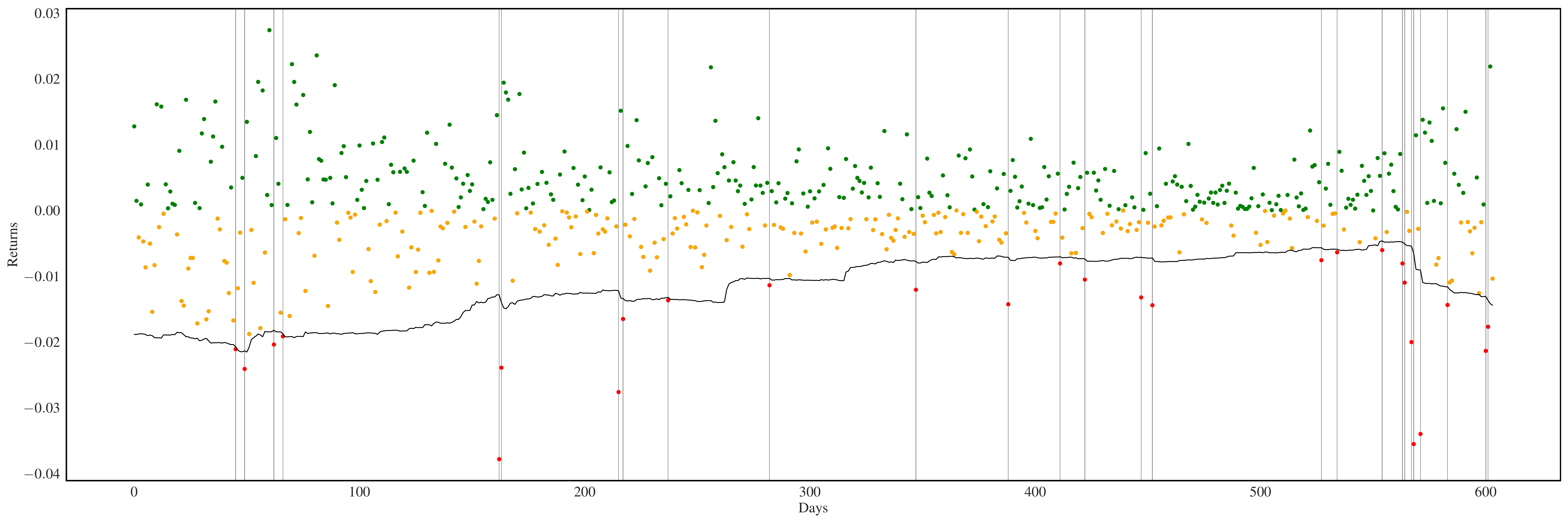}}}\\
    \subfigure[Monte Carlo]{{\includegraphics[width=7.7cm, height=2.1cm]{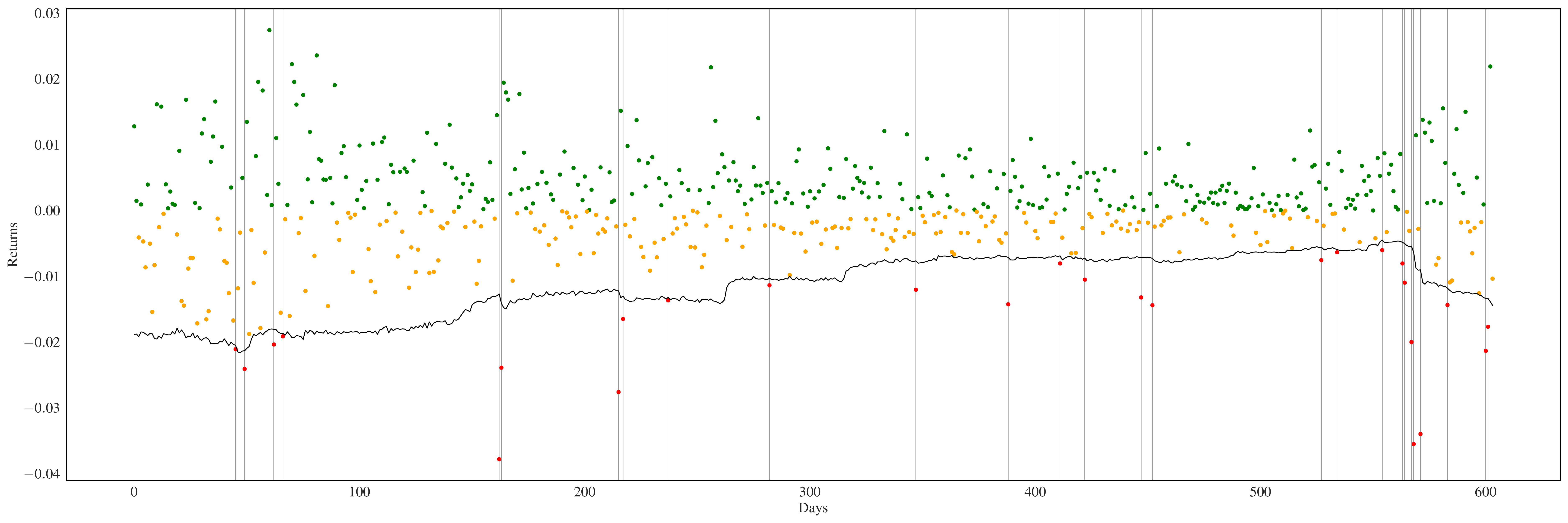}}}
    \subfigure[Encoded VaR]{{\includegraphics[width=7.7cm, height=2.1cm]{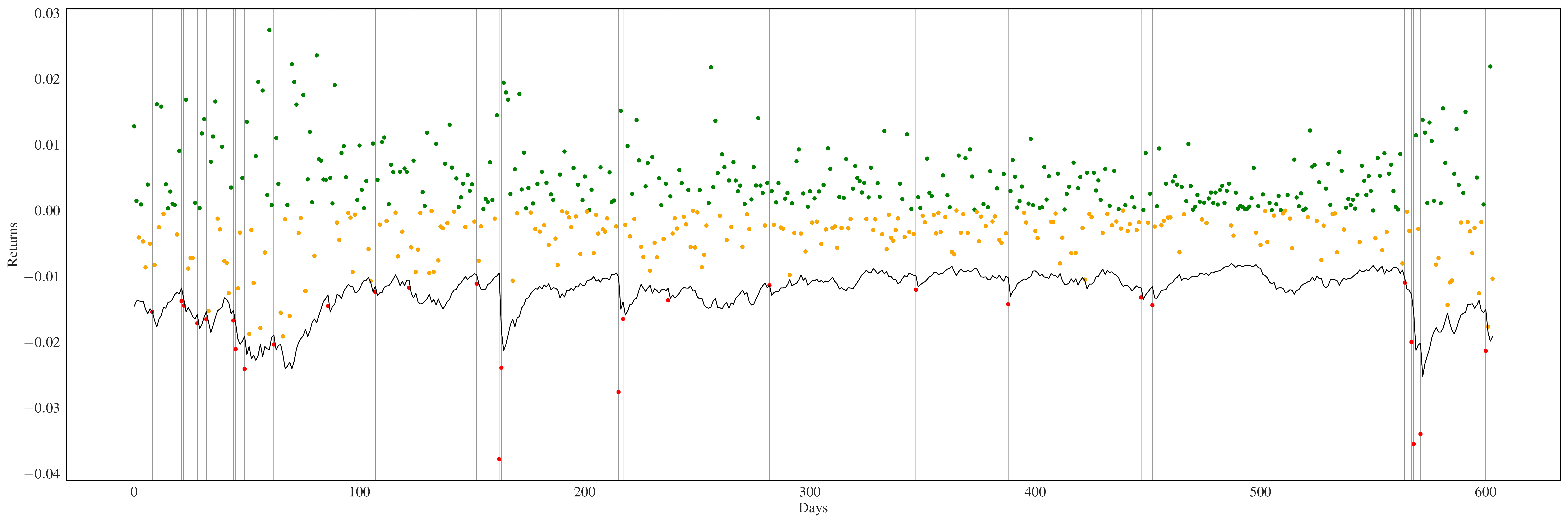}}}
    \caption{Different VaR results for S\&P500 portfolio (testing period). The red dots and the black vertical lines represents the days which the VaR estimation is larger than the actual return.}
    \label{fig:label}
\end{figure}
\section{Conclusions}\label{sec:con}

In this article, we proposed a novel approach to generate cross-section data for portfolio returns. By using the VAE model, we were able to generate data that contained most of the information in the real data. Using a generative model also helped us produce the distribution of portfolio returns in a non-parametric way and estimate VaR.  

Our Encoded VaR methodology has a number of advantages. First, Encoded VaR is a non-parametric approach therefore enforces little technical assumptions for risk forecasting. Second, by re-sampling a simple normal distribution with mean $0$ and standard deviation $1$, it is able to reproduce complex unknown distributions with fat tail properties. Third, Encoded VaR engine is able to decrease the noise that is present in financial data and therefore can increase the signal-to-noise ratio. This is mainly done by using various samples from latent space and regenerating financial scenarios which are similar to historical time series but not the same. Fourth, Encoded VaR come with a Monte Carlo engine which is highly efficient in terms of computational expense as samples from normal distribution. Fifth, our VAE-based model employs a standardization technique which facilitates the learning process of the network and lets the risk manager use their own preferred forecast of the mean and standard deviation. Therefore, Encoded VaR gives risk managers the freedom and flexibility to choose their own preferred expected mean and expected standard deviation, and learns the dependency structure of the data by itself. Sixth, unlike many other risk measurement models, Encoded VaR does not use correlation as a dependency measure for stocks returns. 

A disadvantage of the Encoded VaR, like many other deep learning networks, is that it is not economically interpretable as it leans the return distribution through a complex nonlinear artificial network. Second, as we have seen, Encoded VaR is not ideal to use for regulatory loss functions as it provides a very small margin compared to the general downside return time series. Third, Encoded VaR has a high number of hyper-parameters on which the final VaR result depend. Despite of the disadvantages of our Encoded VaR engine, we believe that machine learning solutions will provide more degree of freedom for quants and academics to model the complex structure of the market's return distributions.

\subsection*{\textit{Managerial Implications}}
There are multiple steps in calculating Encoded VaR that have randomness and might affect the output of the model. These steps include weights initialization of the VAE's neurons, model training, and sampling from latent space for new VaR calculation. We tested the model multiple times, and VaR results showed very low variance. This low variance makes the model applicable.

There are also a number of hyper-parameters which affect the output. Type of regularizer and activation function in layers of the VAE, number of hidden layers, neurons in each hidden layer, reconstruction coefficient in loss function, latent space dimension, and size of the window used for standardization are such hyper-parameters which can be optimized according to the use case. By defining a goodness criterion for the models output, finding these hyper-parameters can be easily automated. 

\subsection*{\textit{Future Research}}
Our methodology for forecasting VaR with generative models can be extended or improved in a number of ways. First, volatility and return prediction models, like ARIMA and GARCH, can be employed along with VAE to forecast the future state of return distributions. Second, other distributions, like Student's t distribution, may be used in the structure of the latent space(\cite{abiri2020variational}). Third, the same VAE engine can be used in other areas of finance and economics where unknown distributions are simulated to forecast returns, risk or portfolio allocation. Fourth, VAEs can be employed for reproducing volatility smiles and smirks in derivatives domains as they can regenerate non-normal features of the options securities. Apart from our VAE-based approach, other deep generative models like Generative Adversarial Networks (GANs) may also be used to reconstruct the return distribution and estimate VaR.


\setcitestyle{numbers} 
\bibliographystyle{plainnat}
\bibliography{refs}

\clearpage

\appendix
\section*{Appendix}
\section{An Overview of Benchmark Models}\label{sec:appendix_benchmark}

In this section, we provide a brief overview of all benchmark methods used in this paper. We divide all VaR methods into four categories: non-parametric methods, parametric methods, semi-parametric methods, and hybrid methods. Non-parametric approaches assume no parametric distribution for the returns and try to extract the distribution from historical data by different techniques. Although these methods have the advantage of not implying any predetermined distribution on the data, they are usually very time consuming. In parametric approaches, simple parametric distributions, like Normal and Student's t are assumed for the returns which speed up computations considerably. However, this speed is reached at the price of ignoring the frequently observed features of time series of returns, like auto-correlation and volatility clustering.  Semi-parametric approaches combine different techniques of parametric and non-parametric approaches. Hybrid methods, on the other hand, employ tools such as time series techniques and combine them with other VaR computation routines. For example, EVT GARCH is a hybrid method of calculating Value-at-Risk based on Extreme Value Theory (EVT).

\subsection{Non-parametric Methods}

\subsubsection{Historical Simulation}
This approach uses a rolling window in historical data and estimates the losses' experimental distribution, then measures the one-period ahead VaR as a predefined quantile of this distribution.

\subsubsection{Monte Carlo Simulation }
This simulation method  simulates future returns based on a predefined stochastic process, then applies a historical simulation method on the simulated data to calculate the next period VaR. In this paper, we have used Geometric Brownian Motion (GBM) for the stock price following the stochastic differential equation
\begin{equation}
    d S_{t}=\mu S_{t} d t+\sigma S_{t} d W_{t},
\end{equation}
where constants $\mu$ and $\sigma$ are called drift and diffusion, respectively, and $W_{t}$ is a Wiener process. 

\subsection{Parametric Methods}

\subsubsection{Variance-Covariance}
First, a rolling window is used and the standard deviation of returns from that window is calculated. Then, assuming normal distribution for returns with mean zero, one can measure VaR at time $t$ using
\begin{equation}
\label{eq:var_covar}
    VaR_{t} = \mathcal{N}^{-1}(\Theta)\sigma_{t},
\end{equation}
where $\mathcal{N}^{-1}$ is the inverse of the cumulative standard normal distribution and $\theta$ is a specific quantile.

\subsubsection{GARCH}

Generalized Autoregressive Conditional Heteroskedasticity (GARCH) is a model to predict future variance using lagged variances and innovations. In GARCH(p,q) model, we have

\begin{equation}
\label{eq_garch}
    \sigma_{t}^2=\alpha_{0}+\sum_{i=1}^{q} \alpha_{i} \varepsilon_{t-i}^{2}+\sum_{j=1}^{p} \beta_{j} \sigma_{t-j}^2,
\end{equation}
where
\begin{equation}
    \varepsilon_{t}=\sqrt{\sigma_{t}} \eta_{t}, \quad \eta_{t}^{\mathrm{IID}} \sim N(0,1),
\end{equation}
$\alpha$ and $\beta$ are constants, and $\sigma_{t}$ is the variance of $\varepsilon_{t}$ conditional on the information available at time $t$. Then, we use equation \ref{eq:var_covar} to calculate VaR. One can use Student's t distribution instead of normal distribution for $\eta_{t}$.

\subsubsection{EGARCH}
EGARCH model is an extension to GARCH model which takes into account the asymmetry of volatility in financial data. In this model, we have
\begin{equation}
    \log \sigma_{t}^{2}=\omega+\sum_{k=1}^{q} \beta_{k} g\left(Z_{t-k}\right)+\sum_{k=1}^{p} \alpha_{k} \log \sigma_{t-k}^{2}
\end{equation}
where $g\left(Z_{t}\right)=\theta Z_{t}+\lambda\left(\left|Z_{t}\right|-E\left(\left|Z_{t}\right|\right)\right)$, $\sigma_{t}^{2}$ is conditional variance, and the coefficients $\omega$, $\beta$, $\alpha$, $\theta$, and $\lambda$ are constant. $Z_{t}$ is a standard normal variable or can be a Student's t variable. 

\subsubsection{RiskMetrics}
This approach is almost similar to the GARCH approach, and it is also called IGARCH (Integrated GARCH) method.
This method uses equation \ref{eq_garch}, with
\begin{equation}
    \sum_{i=1}^{p} \beta_{i}+\sum_{i=1}^{q} \alpha_{i}=1.
\end{equation}
In the case of IGARCH(1,1), $\beta = 0.94$. 

\subsection{Semi-parametric Methods}

\subsubsection{Filtered Historical Simulation}
As a semi-parametric method, Filtered Historical Simulation (FHS), tries to incorporate both characteristics of a non-parametric historical simulation (HS) method, and a parametric method (GARCH). In this method, sampling is done similar to the HS, but after forecasting the volatility using GARCH, the samples are rescaled with the predicted volatility.

\subsubsection{CAViaR Symmetric}
Symmetric  Conditional Autoregressive Value at Risk directly models VaR for return ${x_{t}} $ from previous VaRs of the same series
\begin{equation}
\mathrm{VaR}_{t}=\beta_{1}+\beta_{2} \mathrm{VaR}_{t-1}+\beta_{3}\left|\left(x_{t-1}\right)\right|
\end{equation}

\subsubsection{CAViaR GARCH}
GARCH Conditional Auto-regressive Value at Risk directly models VaR for return ${x_{t}} $ from previous VaRs of the same series such that
\begin{equation}
    \mathrm{VaR}_{t}=\left(\beta_{1}+\beta_{2} \mathrm{VaR}_{t-1}^{2}+\beta_{3}\left(x_{t-1}\right)^{2}\right)^{\frac{1}{2}}
\end{equation}
\subsubsection{CAViaR Adaptive}
Adaptive  Conditional Autoregressive Value at Risk directly models VaR for return ${x_{t}} $ from previous VaRs of the same series such that
$$
\operatorname{VaR}_{t}=\beta_{0}+\sum_{i=1}^{p} \beta_{i} \operatorname{VaR}_{t-i}+\sum_{j=1}^{q} \beta_{j} l\left(x_{t-j}\right)
$$
In this formula, $p$ and $q$ are predefined lags. This method allows for having different coefficients for each return in the series.

\subsubsection{CAViaR asymmetric}
Asymmetric  Conditional Autoregressive Value at Risk directly models VaR for return ${x_{t}} $ from previous VaRs of the same series
\begin{equation}
\mathrm{VaR}_{t}=\beta_{1}+\beta_{2} \mathrm{VaR}_{t-1}+\beta_{3}\left(x_{t-1}\right)^{+}+\beta_{4}\left(x_{t-1}\right)^{-}
\end{equation}
where $\beta_{i}$ are constants, $y^{+} = max(y,0)$ and $y^{-} = - min(y,0)$. The $\beta_{i}$ coefficients are optimized such that the following function is minimized
\begin{equation}
\min _{\beta \in \mathbb{R}^{k}} \frac{1}{T}\left\{(\Theta-1\left(x_{t}<\operatorname{VaR}_{t}\right)\right)\left(x_{t}-\operatorname{VaR}_{t})\right\}
\end{equation}

\section{Sener's Ranking Model}

The ranking model of  \cite{Sener_baronyan_menguturk_2012}, divides the dataset into violation and safe spaces. In the violation space, the magnitude of unexpected losses and unexpected loss clusters are considered. The magnitude of loss in the violation space is defined as the difference between VaR and return
$$ \epsilon_{t}=\left(\mathrm{VaR}_{t}-x_{t}\right) $$
In this ranking model, an unexpected loss cluster is a sequence of successive violations, which is defined as a z-cluster, where z is the number of successive violations in a cluster. A quantity $C_{i}$ is assigned to the cluster number $i$
$$ C_{i}=\prod_{b=1}^{z_{i}}\left(1+\epsilon_{b, i}\right)-1 .$$
For the violation space, penalization measure is the result of interaction between violation clusters. The interaction between clusters $(i)$ and $(i+m)$ is defined as
$$ C_{i} * C_{i+m}=\frac{1}{k_{i, i+m}}\left(\prod_{b=1}^{z_{i}}\left(1+\epsilon_{b, i}\right) \prod_{b=1}^{z_{i+m}}\left(1+\epsilon_{b, i+m}\right)-1\right) $$
Severity of interaction between clusters is defined as the inverse distance between clusters. If the distance between cluster $(i)$ and $(i+m)$ is $k_{i,i+m}$, then the penalization measure for the violation space is defined as
$$ \begin{aligned} \Phi(x, \mathrm{VaR})=& \sum_{i=1}^{\alpha-1} \sum_{m=1}^{\alpha-i} C_{i} * C_{i+m} \\=& \sum_{i=1}^{\alpha-1} \sum_{m=1}^{\alpha-i} \frac{1}{k_{i, i+m}} \\ & \times\left(\prod_{b=1}^{z_{i}}\left(1+\epsilon_{b, i}\right) \prod_{b=1}^{z_{i+m}}\left(1+\epsilon_{b, i+m}\right)-1\right). \end{aligned} $$

Despite the violation space, in the safe space this model only penalizes deviation of VaR from negative returns and does not consider clusters of errors. Penalization measure for the safe space is defined as
$$ \Psi(x, \mathrm{VaR})=\sum_{t=1}^{T}\left[\mathbf{1}\left(x_{t}>\mathrm{VaR}_{t} | x_{t}<0\right)\right]\left(x_{t}-\mathrm{VaR}_{t}\right). $$
The model combines the violation and safe space penalization measures with a weighting parameter which is the selected confidence interval of the underlying VaR model. Therefore, the violation space is weighted with the VaR quantile $\theta$, and the safe space is weighted with $1-\theta$. The total number of observations ($T^{*}$) will be used as a scaling parameter. The combined penalization measure is defined as
$$ \operatorname{PM}(\theta, x, \mathrm{VaR})=\frac{1}{T^{*}}[(1-\theta) \Phi(x, \mathrm{VaR})+\theta \Psi(x, \mathrm{VaR})]. $$
To get a better picture of this measure, \cite{Sener_baronyan_menguturk_2012} calculate the Penalization Measure (PM) ratio as the standardized ratio of the PM value of a certain method with respect to other benchmarks. This ratio for the jth method is defined as
$$ \operatorname{Ratio}_{j}=\frac{\mathrm{PM}_{j}}{\sum_{i=1}^{n} \mathrm{PM}_{i}}. $$
The method with the lowest ratio is the best method and will be ranked first and the methods with higher ratios will be in the next spots.

\end{document}